\documentclass[11pt]{cambrian}
\usepackage[utf8]{inputenc}
\usepackage[
    style=numeric-comp,
    doi=true,
    url=false,
    uniquename=false,
    giveninits=true,
    natbib,
    backend=biber]{biblatex}
\usepackage{colortbl}

\usepackage[utf8]{inputenc} 
\usepackage[T1]{fontenc}    
\usepackage{hyperref}
\usepackage{url}            
\usepackage{booktabs}       
\usepackage{amsfonts}       
\usepackage{nicefrac}       
\usepackage{microtype}      
\usepackage{xcolor}
\definecolor{memoblue}{HTML}{82a2f7}        
\definecolor{memobluedark}{HTML}{5576cc}    
\definecolor{memobluelight}{HTML}{e8eeff}   
\definecolor{bluelink}{RGB}{0,113,188}
\definecolor{greenlink}{RGB}{0,188,113}
\hypersetup{
    colorlinks=true,%
    citecolor=memobluedark,%
    filecolor=memobluedark,%
    linkcolor=black,%
    urlcolor=memobluedark
}
\usepackage{tabularx}
\usepackage[most]{tcolorbox}
\usepackage{amsmath}
\usepackage{multirow}
\usepackage{array}
\usepackage{caption}
\usepackage{wrapfig}
\usepackage{enumitem}
\usepackage{tikz}
\usepackage{lipsum}
\usepackage{subcaption}
\usepackage{multirow} 
\usepackage{adjustbox}

\usepackage{amssymb}
\usepackage[capitalize]{cleveref} 

\usepackage{pgf}
\usepackage{colortbl}

\usepackage{tcolorbox}
\usepackage{rotating}
\usepackage[abs]{overpic}
\usepackage{makecell}
\usepackage{longtable}
\usepackage{algorithm}
\usepackage{algpseudocode}  

\usepackage{tocloft}  

\usepackage[english]{babel}
\usepackage{csquotes}
\usepackage{listings}
\definecolor{codekeyword}{rgb}{0.0, 0.0, 0.5}   
\definecolor{codecomment}{rgb}{0.0, 0.5, 0.0}   
\definecolor{codestring}{rgb}{0.56, 0.0, 1.0}   

\lstdefinestyle{pythonstyle}{
    language=Python,                          
    basicstyle=\ttfamily\small,               
    keywordstyle=\color{codekeyword}\bfseries,
    commentstyle=\color{codecomment}\itshape, 
    stringstyle=\color{codestring},           
    showstringspaces=false,                   
    breaklines=true,                          
    tabsize=4,                                
    numbers=none,                             
    frame=none,                               
    backgroundcolor=\color{white},            
    captionpos=b,                             
    morekeywords={self, __init__, __name__, __main__}, 
}

\usepackage{fontawesome}
\usepackage{xspace}

\usepackage{float}

\usepackage{colortbl}

\newcommand{\NAME}{MEMO}

\correspondingauthor={$^*$Equal Contribution. \quad $^\ddagger$Project Leader. \quad $^\dagger$Equal Advising.}

\title{\quad\\ \vspace*{-1.8cm} \center MEMO: Memory-Augmented Model Context Optimization \\ for Robust Multi-Turn Multi-Agent LLM Games\quad\\ \vspace*{-0.3cm} }

\author{Yunfei~Xie$^{*,1}$, Kevin~Wang$^{*,\ddagger,2}$, Bobby~Cheng$^{*,4}$, Jianzhu~Yao$^{3}$, Zhizhou~Sha$^{2}$, Alexander~Duffy$^{5}$, Yihan~Xi$^{2}$, Hongyuan~Mei$^{6}$, Cheston~Tan$^{4}$, 
Chen~Wei$^{\dagger,1}$, Pramod~Viswanath$^{\dagger,3}$, Zhangyang~Wang$^{\dagger,2}$\\
\small{\textit{$^1$Rice University} \quad \textit{$^2$The University of Texas at Austin} \quad \textit{$^3$Princeton University}}\\
\small{\textit{$^4$A*STAR} \quad \textit{$^5$Good Start Labs} \quad \textit{$^6$TTIC}}
}

\keywords{LLM Agents, Multi-Agent Games, Self-Play, Prompt Optimization, Memory}

\makeatletter
\renewcommand{\abscontent}{%
    \noindent
    \begin{tcolorbox}[
        colback=memobluelight,
        colframe=memobluedark,
        arc=5pt,
        boxrule=0pt,
        left=10pt, right=10pt,
        top=8pt, bottom=8pt,
        enhanced jigsaw,
    ]
    \parbox{\dimexpr\linewidth}{{\fontsize{10}{12}\selectfont \theabstract}}
    \vskip0.4em
    \@ifundefined{@keywords}{}{%
        \noindent{\bfseries\footnotesize\color{memobluedark} Keywords:}~{\footnotesize \@keywords}\par}
    \vskip0.2em
    \noindent{\bfseries\footnotesize\color{memobluedark} Website:}~{\footnotesize\url{https://yunfeixie233.github.io/MEMO}}\\
    {\bfseries\footnotesize\color{memobluedark} Code:}~{\footnotesize\url{https://github.com/openverse-ai/MEMO}}
    \end{tcolorbox}%
}
\makeatother

\begin{document}
\begin{abstract}
Multi-turn, multi-agent LLM game evaluations often exhibit substantial run-to-run variance. In long-horizon interactions, small early deviations compound across turns and are amplified by multi-agent coupling. This biases win rate estimates and makes rankings unreliable across repeated tournaments. Prompt choice worsens this further by producing different effective policies. We address both instability and underperformance with \textbf{\NAME{}} (\textbf{Me}mory-augmented \textbf{MO}del context optimization), a self-play framework that optimizes inference-time context by coupling \textbf{retention} and \textbf{exploration}. Retention maintains a persistent memory bank that stores structured insights from self-play trajectories and injects them as priors during later play. Exploration runs tournament-style prompt evolution with uncertainty-aware selection via \textsc{TrueSkill}, and uses prioritized replay to revisit rare and decisive states. Across five text-based games, \NAME{} raises mean win rate from \textbf{25.1\%} to \textbf{49.5\%} for GPT-4o-mini and from \textbf{20.9\%} to \textbf{44.3\%} for Qwen-2.5-7B-Instruct, using $2,000$ self-play games per task. Run-to-run variance also drops, giving more stable rankings across prompt variations. These results suggest that multi-agent LLM game performance and robustness have substantial room for improvement through context optimization. \NAME{} achieves the largest gains in negotiation and imperfect-information games, while RL remains more effective in perfect-information settings.
\end{abstract}

\maketitle
\vspace{0.1cm}

\begin{figure}[H]
    \centering
    \vspace{-3px}
    \begin{subfigure}[b]{0.49\linewidth}
        \centering
        \includegraphics[width=.95\linewidth, keepaspectratio=false]{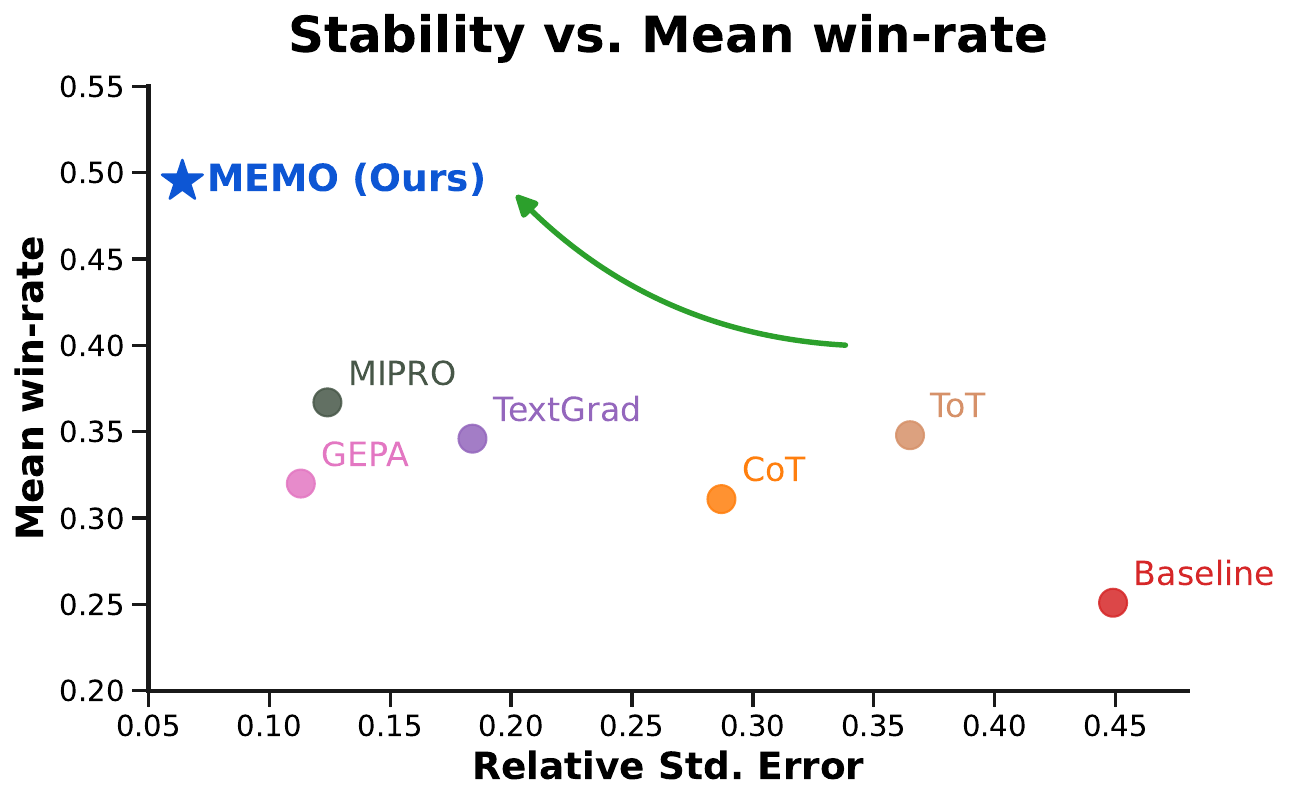}
        \caption{\textbf{Performance and stability across methods.}}
        \label{fig:teaser_a}
    \end{subfigure}
    \vspace{-3px}
    \hfill
    \begin{subfigure}[b]{0.48\linewidth}
        \centering
        \includegraphics[width=.9\linewidth, keepaspectratio=false]{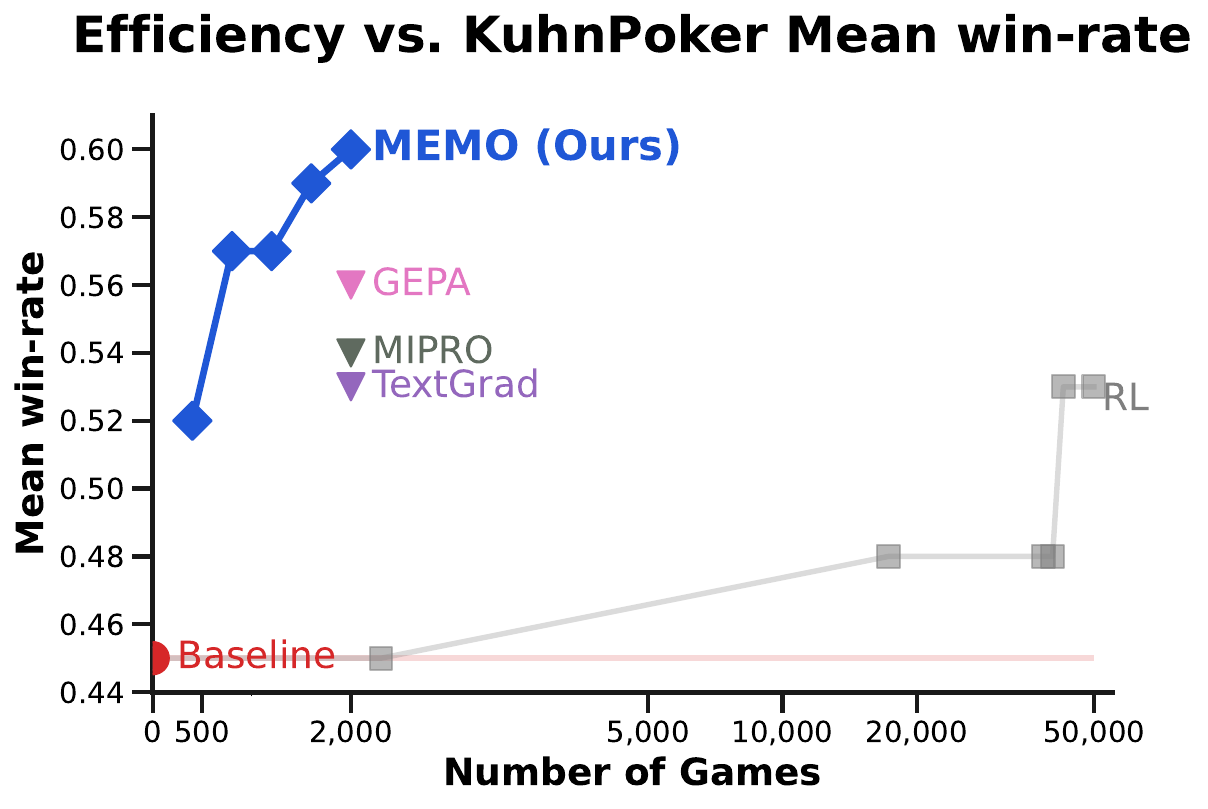}
        \caption{\textbf{Training Efficiencies in \textsc{KuhnPoker}.}}
        \label{fig:teaser_b}
    \end{subfigure}
    \vspace{-3px}
    \label{fig:teaser}
\caption{\textbf{Left} Run-to-run performance and stability comparison. Using GPT-4o-mini with \NAME{} achieves the highest mean win rate (49.5\%) with the lowest RSE (6.4\%). \textbf{Right} Learning efficiency comparison against the self-play RL baseline method \textsc{Unstablebaseline}. Using Qwen2.5-7B-Instruct, \NAME{} reaches 60\% win rate on Kuhn Poker with only 2{,}000 games, 19$\times$ fewer than the 38{,}000 games required by the RL self-play baseline.}
\end{figure}


\section{Introduction}

Large language models (LLMs) have rapidly saturated many static benchmarks, leaving limited headroom for single-turn QA and reasoning datasets such as AIME~\citep{aime2024}, SWE-Bench~\citep{jimenez2023swe}, and GPQA~\citep{rein2024gpqa}. This shifts attention toward multi-turn and interactive evaluations, namely game-based benchmarks~\citep{duan2024gtbench,topsakal2024evaluating,fan2024can}, which stress long-horizon reasoning, adaptation, and strategic interaction. Games are easy to simulate, come with objectives, and require capabilities that apply to real-world challenges such as planning under uncertainty, negotiation, and context-sensitive decision making.

However, \emph{multi-turn, multi-agent LLM evaluation is inherently unstable}. Because each model output becomes part of the subsequent input, small early deviations can compound across turns, leading to divergent trajectories~\citep{laban2025llms}. In multi-agent games, interaction coupling can worsen this effect. An inconsistent response from one agent can perturb the other agent's best responses, reshaping the joint trajectory~\citep{cemri2025multi}. Separately, some LLMs exhibit nondeterministic outputs even under nominally deterministic decoding settings~\citep{blair2025llms}. From an evaluation perspective, these factors can bias win-rate estimates and destabilize comparative rankings across repeated tournaments, complicating reproducibility and fair model comparison.

Inference-time \emph{context}, including prompts, instructions, and auxiliary information, offers a direct lever for performance in interactive settings. Small contextual variations can induce different effective policies and rank reversals across models (Appx.~\ref{appendix:prompt_sensitivity}), motivating treatment of context not as a fixed wrapper but as an \emph{agentic object} that should be optimized under interaction.

Existing approaches, however, struggle in multi-turn, path-dependent games. Prompt engineering techniques such as chain-of-thought (CoT)~\citep{wei2022chain} instructions or hand-designed templates remain fixed throughout evaluation. While these can improve win rate or reduce superficial errors, they do not adapt to failure modes or strategic patterns that emerge through interaction. Automatic prompt optimization methods~\citep{yuksekgonul2024textgrad,yin2025llm,agrawal2025gepa,opsahl2024optimizing} allow prompts to adapt, but are largely developed for static tasks. They update prompts using feedback from a local batch of trajectories and lack persistent memory. In multi-turn, multi-agent games, different tournaments surface different decisive states and rare failure modes; without a mechanism to retain and reuse insights across rounds, prompt optimization becomes run-dependent, leading to high variance in both learned contexts and performance.

We therefore propose \textbf{MEMO} (\textbf{Me}mory-augmented \textbf{MO}del context optimization), a self-play framework that optimizes inference-time context without updating model weights. MEMO couples \emph{exploration}, tournament-style context evolution with uncertainty-aware selection via \textsc{TrueSkill} and prioritized replay, with \emph{retention}, a persistent memory bank that distills self-play trajectories into structured insights through create, read, update, and delete (CRUD) style operations and reinjects them as priors in subsequent rounds. The central finding is that exploration alone yields only modest gains; persistent memory is what transforms context optimization from a memoryless search into a cumulative learning process.

Across five text-based games from \texttt{TextArena} and \texttt{SPIN-Bench}~\citep{guertler2025textarena,yao2025spin}, \NAME{} raises mean win rate from \textbf{25.1\%} to \textbf{49.5\%} for \textbf{GPT-4o-mini}~\citep{openai2024gpt4o_mini} and from \textbf{20.9\%} to \textbf{44.3\%} for \textbf{Qwen-2.5-7B-Instruct}~\citep{yang2025qwen2_5}. It uses only 2{,}000 self-play games per task, 19$\times$ fewer than RL baselines, while reducing run-to-run variance by 7$\times$ to a relative standard error of 6.4\% compared to 43.3\%.

We make three main contributions.
\begin{itemize}[leftmargin=1.5em, labelindent=0pt, labelsep=0.5em,topsep=1pt, itemsep=1pt]

  \item \textbf{Context sensitivity in multi-turn, multi-agent LLM games.}
  We show that evaluation outcomes are sensitive to context choices. 
  Small prompt variations can shift effective policies and alter model rankings, motivating robust practices such as prompt-variation reporting rather than reliance on single-prompt evaluations.

  \item \textbf{A unified framework of reflection, memory, and replay.}
  We introduce a framework that combines structured reflection, persistent memory, context evolution, and prioritized replay, allowing the agent to accumulate and reuse knowledge across rounds rather than discarding it at each update.

  \item \textbf{Training-efficiency gains with improved stability.}
  We report that MEMO substantially improves win rates under a fixed self-play budget while reducing run-to-run variance of end-to-end outcomes.
  It achieves competitive or stronger results than existing prompt optimization methods in imperfect information games, while RL remains more effective in perfect-information settings.
\end{itemize}

\begin{figure*}[t]
    \centering
    \includegraphics[width=\textwidth]{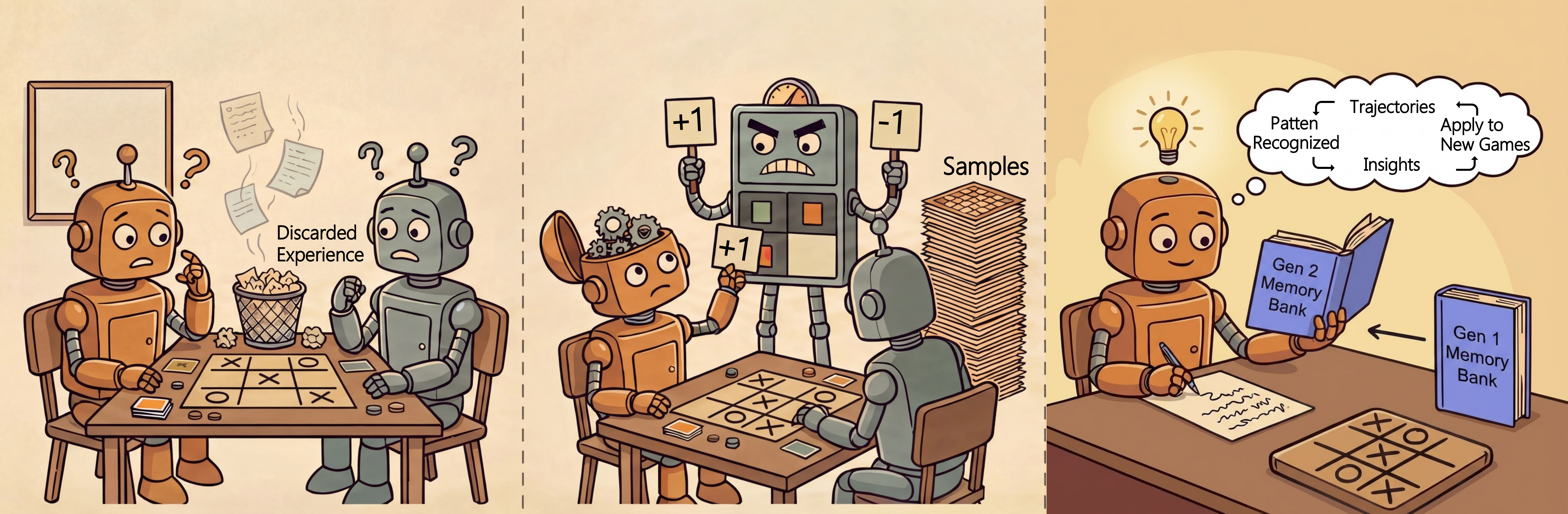}\\[2pt]
    \makebox[0.333\textwidth]{\small\textbf{(a) Self-Play Prompt Optimization}}%
    \makebox[0.333\textwidth]{\small\textbf{(b) Reinforcement Learning}}%
    \makebox[0.333\textwidth]{\small\textbf{(c) \NAME{} (Ours)}}
    \caption{\textbf{Three paradigms for learning in multi-agent LLM games.}
    \textbf{(a)}~Prompt optimization updates the system prompt each round
    through self-play, but game experience is not effectively retained across rounds,
    so strategic insights are lost across rounds.
    \textbf{(b)}~Reinforcement learning (RL) updates model weights through
    self-play but relies on outcome rewards, requiring large sample budgets.
    \textbf{(c)}~\NAME{} reflects on completed trajectories and accumulates
    reusable insights in a persistent memory bank across generations,
    enabling improvement without weight updates or external reward.}
    \label{fig:conceptual}
\end{figure*}




\section{Preliminary and Problem Statement}
\label{problem_statement} 
\paragraph{Two-Player Multi-Turn Markov Game.}
We formalize the setting as a two-player, turn-based, zero-sum, partially observable Markov game $(S, A, O, P, \Omega, \rho)$, where $S$ is the state space, $A$ is the action space where each action is a complete model response, $O$ is the observation space, $P{:}\, S \times A \to \Delta(S)$ governs transitions, $\Omega{:}\, S \to O$ maps states to partial observations, and $\rho{:}\, S_{\text{term}} \to \{-1, 0, 1\}$ assigns win/draw/loss at terminal states. Players alternate turns; a trajectory $\tau = (s_0, a_0, \ldots, s_H)$ terminates after $H$ steps with outcome $r_0(\tau) = \rho(s_H)$ for Player~0.

\paragraph{Prompt and Memory as Game Context.}
We define \emph{context} as all information that conditions the model before and during play. Let $c=(q,M)$, where $q$ is the instruction prompt, including role and system text fixed at game start,
and $M$ is the memory injected at inference time without weight updates. $M$ consists of structured, reusable insights distilled from past self-play trajectories.
In \NAME{}, $M$ is drawn from a persistent memory bank $\mathcal{B}_{\text{mem}}$ that accumulates across
optimization iterations, and each game instance may use a subsampled memory $M\subseteq \mathcal{B}_{\text{mem}}$.

\paragraph{Full-Context Evaluation.}
\label{RSE}
We evaluate each method over $n$ independent runs of its full context-optimization pipeline, each producing a final context $c^*$ that is evaluated on a fixed game suite $\mathcal{G}$. For each game, we play multiple rounds against a fixed opponent pool, swapping first-move order to reduce bias (opponents use the reference contexts in Appx.~\ref{appendix: basepromptexample}). Let $x_r$ denote the run-level performance, defined as the mean win rate averaged over all games, opponents, and rounds. We report the mean performance across runs, $\mathrm{mean}(x_1,\ldots,x_n)$, together with the relative standard error $\mathrm{RSE}(\%) = 100 \times \frac{\mathrm{std}(x_1,\ldots,x_n)}{\mathrm{mean}(x_1,\ldots,x_n)\sqrt{n}}$, where lower RSE indicates greater run-to-run stability.

\begin{figure*}[t]
    \centering
    \includegraphics[width=0.98\linewidth]{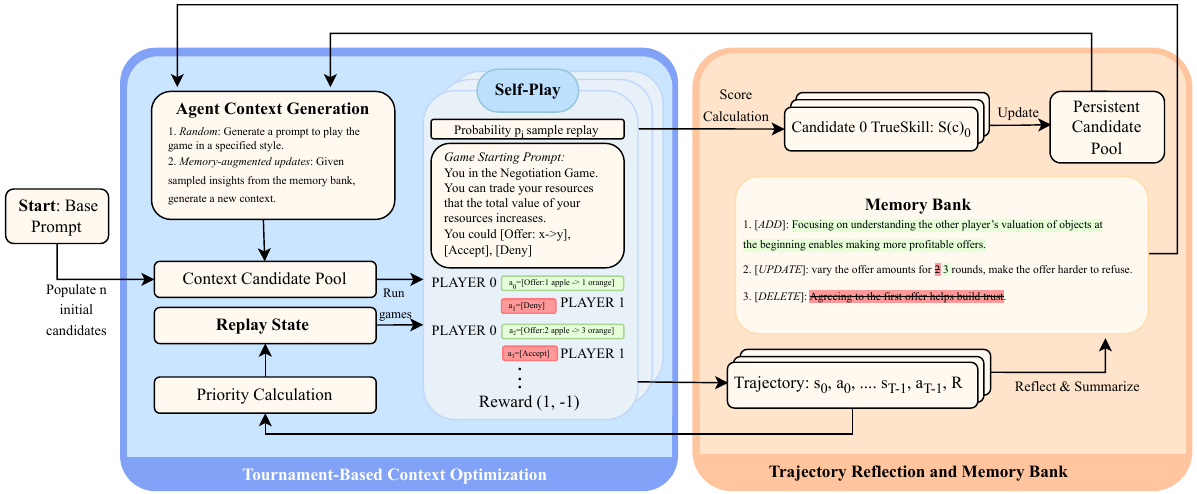}
    \caption{\textbf{The \NAME{} Framework}. At each optimization generation, new candidate contexts are proposed through two strategies: random proposals and memory-augmented updates. These candidates are then evaluated via self-play, and the best-performing candidates are used to update the pool for the next generation. To encourage exploration and mitigate redundant early moves, a prioritized replay module is introduced, enabling efficient search for robust prompts and priors within a single game.}
    \label{fig:methodology}
\end{figure*}

\section{The \NAME{} Framework}
\label{method}

\NAME{} operates over multiple optimization generations. Each generation $g$ consists of a self-play tournament, context evolution (Sec.~\ref{sec:prompt_optimization}), insight extraction from trajectories (Sec.~\ref{sec:memory}), and state selection for replay (Sec.~\ref{sec:replay}). Fig.~\ref{fig:methodology} provides an overview and Appx.~\ref{appendix:ablation} details hyperparameter tuning.

\subsection{Tournament-Based Context Optimization}\label{sec:prompt_optimization}


\paragraph{Context selection via game outcomes.}
 \NAME{} maintains a population of $N$ candidate contexts, each defining a different prompt and set of priors for the agent. The core idea is to evaluate each of these candidate context by its game performance so that contexts which lead to wins are retained for the next generation, while those which result in losses are discarded. Let $\mathcal{C}_g$ denote the \emph{context population} at optimization generation $g$. Each context $c\in\mathcal{C}_g$ is evaluated via multi-agent self-play in games against a baseline agent, the same base model using only a default prompt; see Appx.~\ref{appendix: basepromptexample}. For asymmetric games, each round consists of two games with roles swapped to remove first-move bias. These matches produce win/loss outcomes for each context, but raw win counts are unreliable when games are limited. A context that wins 3 out of 3 games may simply be lucky rather than genuinely strong. To address this, we use \textsc{TrueSkill}~\citep{herbrich2006trueskill}, a Bayesian skill rating that models each context's skill as a Gaussian with mean $\mu_c$ and uncertainty $\sigma_c$. We select contexts using a conservative lower-confidence bound: \begin{equation} \label{equation:TrueSKill} S(c)=\mu_c-\kappa\,\sigma_c, \end{equation} where $\kappa$ is a penalty coefficient (see Sec.~\ref{sec:hyperparamselection}). This penalizes contexts with high uncertainty, favoring those that win reliably across multiple observations.

\paragraph{Context generation for the next generation.}

After selection, low-scoring contexts are discarded, leaving the population incomplete. To restore the population to size $N$ for the next generation, we generate new candidate contexts. Across optimization generations, we maintain a \emph{persistent candidate pool} $\mathcal{P}$ that stores the best contexts observed so far. After evaluating the current population $\mathcal{C}_g$, we update $\mathcal{P}$ by retaining only the top-scoring candidates from $\mathcal{P}\cup\mathcal{C}_g$. We then form the next generation's population $\mathcal{C}_{g+1}$ using two proposal operators, where a fraction of new candidates are generated via \textit{random proposals} and the remainder via \textit{memory-augmented updates}; see Sec.~\ref{sec:hyperparamselection} for the specific ratio.


\begin{enumerate}[leftmargin=1.5em, labelindent=0pt, labelsep=0.5em,topsep=1pt, itemsep=1pt] \item \textbf{Random proposals.} Introduce novel variations to encourage exploration by sampling a playstyle from a fixed catalog and applying small, length-bounded edits to the base context to instantiate that style while preserving legality and interface constraints (Appx.~\ref{appendix:prompt-ops-random}). \item \textbf{Memory-augmented updates.} Incorporate insights extracted from trajectory reflections (Sec.~\ref{sec:memory}) into targeted prompt edits. \end{enumerate} Note that in the first generation ($g=0$), the memory bank is empty, so all initial contexts are generated via random proposals. 

After the final optimization generation, \NAME{} outputs the highest-scoring context in \(\mathcal{P}\): \[ C^\star = \arg\max_{C\in\mathcal{P}} S(C). \]

\subsection{Trajectory Reflection and Memory Bank}
\label{sec:memory}

This section describes the \emph{retention} component of \NAME{}, which preserves and combines insights across optimization generations. Multi-turn games make post-hoc attribution easier than online decision making because a completed trajectory reveals which choices led to the observed outcome, relating to hindsight-style analysis~\citep{andrychowicz2017hindsight}. \NAME{} exploits this by extracting structured insights from completed self-play trajectories and storing them in a persistent memory bank.

\paragraph{Trajectory reflection.}
After each optimization generation, we sample a fixed number of completed self-play trajectories and prompt the model to extract a small set of typed insights, e.g., rule clarifications, legality constraints, and strategy priors. 
For each sampled trajectory, the model reviews the sequence of states, actions, and final outcome, then produces one or more candidate insights that summarize lessons learned. 
These insights capture what worked, what failed, and why, providing structured feedback that can inform future play. 
The reflection prompt template is provided in Appx.~\ref{appendix:reflection-prompt}.

\paragraph{Memory bank.}
\NAME{} maintains a shared memory bank \(\mathcal{B}_{\text{mem}}\) that persists across optimization generations.
For each generation with $N$ evaluated trajectories, the reflection step produces up to $N$ candidate insights that must be reconciled with the existing memory bank.
Following database-style operations~\citep{Martin1983ManagingDBEnv}, we merge new insights into \(\mathcal{B}_{\text{mem}}\) using three operations.
\begin{enumerate}[leftmargin=1.5em, labelindent=0pt, labelsep=0.5em,topsep=1pt, itemsep=1pt]
\item \textbf{Add.} If a new insight is not similar to any existing insight in the memory bank, it is added directly.
\item \textbf{Remove.} If a new insight conflicts with an existing insight, meaning they suggest contradictory strategies or conclusions, both the new and existing insights are removed to avoid misleading the agent.
\item \textbf{Edit.} If a new insight is similar to an existing one, the two are merged by enhancing, generalizing, or improving the existing insight to be more actionable.
\end{enumerate}
The agent compares each candidate insight against the current memory bank and applies the appropriate operation. This merge procedure allows the memory bank to grow, refine, and self-correct over time. The memory operation prompt is provided in Appx.~\ref{appendix:memory-operation-prompt}.

In the next optimization generation, we sample a compact subset \(M \subseteq \mathcal{B}_{\text{mem}}\) and append it to the context of a fraction $\pi$ of the candidate population during self-play, where $\pi$ controls what proportion of agents receive memory-based initialization. This provides reusable, game-specific priors at inference time; see Sec.~\ref{sec:hyperparamselection} for specific values. 
The same memory bank also conditions the memory-augmented proposal operator, enabling targeted prompt edits that reuse aggregated lessons rather than relying only on the most recent tournament.

\subsection{Prioritized Replay}
\label{sec:replay}
Trajectory reflection improves retention, but exploration alone does not guarantee that rare or decisive states will be revisited. To improve trajectory coverage, \NAME{} maintains a replay buffer $\mathcal{B}_{\text{rep}}$ that stores trajectory prefixes together with the environment seed needed to reproduce them. Because storage occurs at each turn within an episode, replayed trajectories need not cover a full game. Invalid moves are retained to preserve the \textit{unaltered course of play}, ensuring that replays faithfully reflect the original gameplay dynamics. To avoid dominance by common action patterns, the buffer \textit{biases sampling toward infrequently encountered trajectories}, encouraging a more diverse and balanced pool of prompt-level insights. We prioritize rare prefixes using an inverse-frequency score, defined for a stored prefix $\tau$ as $\mathrm{priority}(\tau)=\tfrac{1}{\mathrm{count}(\tau)}$. During sampling, the probability $p_i$ of selecting trajectory $\tau_i$ is obtained by raising its priority to a power $\alpha>0$ and normalizing over the buffer,
$p_i = \tfrac{\mathrm{priority}(\tau_i)^{\alpha}}
            {\sum_{j=1}^{|\mathcal{B}_{\text{rep}}|}
             \mathrm{priority}(\tau_j)^{\alpha}}$,
where $|\mathcal{B}_{\text{rep}}|$ denotes the current number of stored trajectories.

The buffer is first populated during generation~0 and becomes available from generation~1 onward.
A gating parameter~$\beta$, the replay probability, determines how often games are initialized from the replay buffer rather than played afresh. 
When replay is chosen, the stored trajectory prefix, that is, the sequence of past player actions, corresponding game states, and the associated game's random seed, is injected into the environment, ensuring faithful reproductions of past episodes while balancing new exploration. Specific values for $\alpha$, $\beta$, and buffer capacity $B$ are provided in Sec.~\ref{sec:hyperparamselection}.

\section{Experiment Setup}
\label{experiment}

\subsection{Game Environments}
Following prior interactive evaluation suites such as LMGame-Bench and BALROG~\citep{hu2025lmgamebenchgoodllmsplaying, paglieri2025balrogbenchmarkingagenticllm}, our games span core problem classes studied in game theory and multi-agent systems. We group them into three categories. \textbf{Negotiation} games, which test cooperation and compromise~\citep{negotiationandhonesty, abdelnabi2024llmdeliberation}; \textbf{Imperfect Information} games, which require reasoning under uncertainty and partial observability~\citep{DBLP:journals/corr/abs-2007-13544, guo2024suspicionagent}; and \textbf{Perfect Information} games, which emphasize planning and long-horizon decision-making with full state visibility~\citep{DBLP:journals/corr/abs-1712-01815}. See Appx.~\ref{app:game_env} for environment descriptions.

\subsection{Baselines and Evaluation Protocol}
\label{sec:optmization_settings}
\label{evaluation_setup}

We compare \NAME{} against three classes of methods.
\textbf{Static prompting} uses unoptimized contexts, including the default \texttt{TextArena} prompt as a baseline, chain-of-thought (CoT), and tree-of-thought (ToT). The baseline prompt is shown in Appx.~\ref{appendix: basepromptexample}.
\textbf{Prompt optimization} adapts the context through feedback, including \texttt{TextGrad}~\citep{yuksekgonul2024textgrad}, \texttt{MIPRO}~\citep{opsahl2024optimizing}, and \texttt{GEPA}~\citep{agrawal2025gepa}.
\textbf{RL} updates model weights through self-play, including \texttt{UnstableBaselines}~\citep{Guertler_UnstableBaselines_2025} and \texttt{SPIRAL}~\citep{liu2025spiral}.
Configurations for all methods are provided in Appx.~\ref{appendix: Training_settings}.

All experiments use \textbf{GPT-4o-mini}~\citep{openai2024gpt4o_mini} and \textbf{Qwen-2.5-7B-Instruct}~\citep{yang2025qwen2_5} as base models.
For prompt-based methods, we perform three independent optimization runs; each resulting context is evaluated against held-out opponents (Grok-4-Fast-Non-Reasoning~\citep{grok4_fast_nonreasoning_2025}, Gemini-2.5-Flash-Lite~\citep{comanici2025gemini}, and Qwen3-235B-A22B-Instruct-2507~\citep{yang2025qwen2_5}) over 50 games per opponent per run.
For RL methods, we train a single policy, select the best checkpoint, and evaluate over three sets of 50 games against the same opponents.
We report mean win rates and relative standard error (RSE; defined in Sec.~\ref{RSE}) across runs. A fixed sampling temperature of $\tau=1.0$ is used throughout.

\subsection{Hyperparameter Selection}
\label{sec:hyperparamselection}
We use a single, fixed configuration across all experiments to avoid per-task tuning; ablation results are in Appx.~\ref{appendix:ablation}. 

\textbf{Context optimization loop.}
We maintain a population of $N=8$ candidate contexts and run $G=5$ optimization generations.
In each generation, we collect $S=50$ self-play games per candidate (total $N \times G \times S = 2000$ games).
We set the TrueSkill penalty coefficient to $\kappa = 1$.

\textbf{Memory-augmented initialization.} We control what proportion of the candidate population receives insights from the shared memory bank $\mathcal{B}_{\text{mem}}$ at initialization. We denote this proportion by $\pi \in [0,1]$, where $\pi=0$ means no candidates receive memory and $\pi=1$ means all candidates are initialized with sampled insights. We use $\pi=0.75$.

\textbf{Replay mechanism.} The replay mechanism uses three hyperparameters. Buffer capacity $B$ sets the maximum number of stored trajectories. Priority exponent $\alpha$ controls the strength of prioritizing rare trajectories. Replay gate $\beta$ sets the probability of initializing from replay rather than starting a new game. We use $B=100{,}000$, $\alpha=0.6$, and $\beta=0.4$.
\section{Results and Analysis}

\begin{wraptable}{r}{0.52\columnwidth}
    \centering
    \vspace{-0.8em}
    \setlength{\tabcolsep}{2.5pt}
    \begin{tabular}{@{}lrrrr@{}}
    \toprule
    \textbf{Optimizer} & \textbf{SimpNeg} & \textbf{Kuhn} & \textbf{SimpTak} & \textbf{Avg.} \\
    \midrule
    TextGrad              &       842 &       986 &       938 &       922 \\
    MIPRO                 & 145{,}864 & 162{,}084 & 754{,}534 & 354{,}161 \\
    GEPA                  & 110{,}325 & 119{,}365 & 111{,}907 & 113{,}865 \\
    \textbf{\NAME{}} &  87{,}364 &  94{,}160 &  89{,}152 &  90{,}575 \\
    \bottomrule
    \end{tabular}
    \caption{Output token cost per prompt optimization method across three games.}
    \label{tab:token-cost}
    \vspace{-1em}
\end{wraptable}

\paragraph{Observation 1. Persistent self-play memory enables sample-efficient and stable gains.}
As shown in Tab.~\ref{tab:main-results_v2}, \NAME{} consistently outperforms other prompt optimization methods, achieving an average gain over \texttt{TextGrad} (14.9\%), \texttt{MIPRO} (12.8\%), and \texttt{GEPA} (17.5\%) with GPT-4o-mini. While the margin relative to RL-based methods such as \texttt{UnstableBaselines} and \texttt{SPIRAL} is smaller, \NAME{} remains competitive while using \textbf{19$\times$ fewer} environment interactions (\textbf{2{,}000} vs.\ \textbf{38{,}000} games).

\definecolor{StaticGroup}{RGB}{232,245,236}

\definecolor{OptGroup}{RGB}{251,242,230}

\definecolor{RLGroup}{RGB}{247,233,229}

\definecolor{OursBlue}{RGB}{220,235,255}

\definecolor{BaselineCell}{RGB}{244,210,210}

\begin{table*}[t]
\centering
\resizebox{0.95\textwidth}{!}{
\begin{tabular}{l l c c c c c c c}
\toprule
\multirow{2}{*}{\textbf{Type}} & \multirow{2}{*}{\textbf{Optimizer}} & \multicolumn{2}{c}{\textbf{Negotiation}} & \multicolumn{2}{c}{\textbf{Imperfect Info}} & \multicolumn{1}{c}{\textbf{Perfect Info}} & \multirow{2}{*}{\begin{tabular}{@{}c@{}}\textbf{Mean}\\\textbf{Win Rate}\end{tabular}} & \multirow{2}{*}{\begin{tabular}{@{}c@{}}\textbf{Mean}\\\textbf{RSE}\end{tabular}} \\
\cmidrule(lr){3-4}\cmidrule(lr){5-6}\cmidrule(lr){7-7}
& & \textbf{SimpleNegotiation} & \textbf{TwoDollar} & \textbf{KuhnPoker} & \textbf{Briscola} & \textbf{SimpleTak} & & \\
\midrule

\multicolumn{9}{l}{\textbf{GPT-4o-mini}} \\
\addlinespace[0.3em]

\rowcolor{StaticGroup}
\textsc{Static} & \textbf{baseline} & 31.3\% & 32.2\% & 39.1\% & 0.3\% & 21.4\% & 25.1\% & 44.9\% \\
\rowcolor{StaticGroup}
& CoT & 27.8\% & 25.7\% & 46.5\% & 30.4\% & 24.8\% & 31.1\% & 28.7\% \\
\rowcolor{StaticGroup}
& ToT & 26.3\% & 27.0\% & 51.7\% & 45.1\% & 23.8\% & 34.8\% & 36.5\% \\

\rowcolor{OptGroup}
\textsc{Prompt} & TextGrad & 42.0\% & 44.6\% & \textbf{55.6}\% & 7.1\% & 23.6\% & 34.6\% & 18.4\% \\
\rowcolor{OptGroup}
& MIPRO & 38.4\% & 50.9\% & 55.1\% & 19.7\% & 19.1\% & 36.7\% & 12.4\% \\
\rowcolor{OptGroup}
& GEPA & 36.8\% & 40.4\% & 52.2\% & 3.3\% & 26.9\% & 32.0\% & 11.3\% \\

\rowcolor{OursBlue}
\textsc{Ours} & \textbf{MEMO} & \textbf{54.9\%} & \textbf{52.4\%} & \textbf{55.6\%} & \textbf{42.7\%} & \textbf{41.8\%} & \textbf{49.5\%} & \textbf{6.4\%} \\

\midrule
\multicolumn{9}{l}{\textbf{Qwen2.5-7B-Instruct}} \\
\addlinespace[0.3em]

\rowcolor{StaticGroup}
\textsc{Static} & \textbf{baseline} & 24.0\% & 17.1\% & 45.3\% & 2.8\% & 15.1\% & 20.9\% & 30.1\% \\
\rowcolor{StaticGroup}
& CoT & 23.8\% & 18.7\% & 42.0\% & 25.8\% & 13.6\% & 24.8\% & 43.4\% \\
\rowcolor{StaticGroup}
& ToT & 27.1\% & 20.7\% & 42.2\% & 22.7\% & 15.1\% & 25.6\% & 40.2\% \\

\rowcolor{OptGroup}
\textsc{Prompt} & TextGrad & 37.1\% & 29.3\% & 52.8\% & 7.1\% & 22.4\% & 29.9\% & 21.7\% \\
\rowcolor{OptGroup}
& MIPRO & 42.4\% & 47.5\% & 53.8\% & 2.2\% & 20.9\% & 33.4\% & 7.3\% \\
\rowcolor{OptGroup}
& GEPA & 34.4\% & 31.7\% & 55.8\% & 3.3\% & 19.3\% & 28.8\% & 14.8\% \\

\rowcolor{RLGroup}
\textsc{RL} & UnstableBaseline & 41.1\% & 30.4\% & 52.7\% & \textbf{53.3\%} & \textbf{47.3\%} & \textbf{45.0\%} & 43.3\% \\
\rowcolor{RLGroup}
& SPIRAL & 45.7\% & -- & 56.7\% & -- & 32.7\% & -- & -- \\

\rowcolor{OursBlue}
\textsc{Ours} & \textbf{MEMO} & \textbf{48.0\%} & \textbf{48.4\%} & \textbf{60.0\%} & 31.1\% & 34.0\% & \textbf{44.3\%} & \textbf{6.1\%} \\

\bottomrule
\end{tabular}
}
\caption{Benchmark results for different approaches using GPT-4o-mini and Qwen2.5-7B-Instruct across multiple tasks. Each win rate is the mean across three evaluation models. \textbf{Type} denotes the optimization paradigm: Static prompting, Prompt optimization, Reinforcement learning (RL), and our method. For full model-wise results, see Appendix~H.}
\label{tab:main-results_v2}
\end{table*}

\textbf{Sample-efficient gains.}
These gains stem from \NAME{}'s ability to accumulate reusable, game-specific insights in the persistent memory bank across self-play episodes (Fig.~\ref{fig:teaser_b}). Qualitative analysis of stored insights (Appx.~\ref{appendix:insightcaseanalysis}) reveals that high-quality entries encode transferable strategic principles rather than instance-specific action reminders. In \textsc{KuhnPoker}, the memory bank learns pressure-based betting heuristics that balance aggression with hand strength. In \textsc{SimpleNegotiation}, it discovers that opponents hold asymmetric resource valuations, a concept never stated in the game rules, and learns to probe preferences before committing to offers. In \textsc{TwoDollar}, it captures time-pressure tactics that exploit the finite round structure. These abstractions persist across optimization generations while less informative or overly specific feedback is gradually diluted through the memory merge operations (Sec.~\ref{sec:memory}). Unlike prompt-only optimization methods that reset context after each update, \NAME{} retains and compounds information across generations, allowing performance improvements to accumulate with substantially fewer interactions.

Retaining high-value insights also improves computational efficiency. As shown in Tab.~\ref{tab:token-cost}, \NAME{} uses only 91K output tokens on average, about one-quarter of \texttt{MIPRO} (354K) and 20\% fewer than \texttt{GEPA} (113K), while achieving similar or better win rates (Tab.~\ref{tab:main-results_v2}). Methods such as \texttt{MIPRO} and \texttt{GEPA} rely on many reflective rollouts and prompt revisions, increasing token usage without commensurate performance gains, while \texttt{TextGrad} uses very few tokens ($\sim$1K) but lacks capacity to learn complex multi-turn behaviors. By retaining high-value insights and reusing them across generations, \NAME{} concentrates learning on fewer, more informative interactions, improving the trade-off between token cost, interaction budget, and win rate.
\begin{table}[t]
    \centering
    \vspace{-0.5em}

    \scriptsize
    \setlength{\tabcolsep}{0.13cm}
    \definecolor{expgrey}{RGB}{245,248,252}

    \resizebox{0.98\textwidth}{!}{
    \begin{tabular}{ccc|ccc|cc}
    \toprule
    \multicolumn{3}{c|}{\textbf{Modules}} &
    \multicolumn{3}{c|}{\textbf{Win Rate}} &
    \multicolumn{2}{c}{\textbf{Summary}} \\

    \textbf{Tournament} & \textbf{Mem} & \textbf{Replay}
    & \textbf{TwoDollar} & \textbf{KuhnPoker} & \textbf{Briscola}
    & \textbf{Mean} & $\boldsymbol{\Delta_{\text{base}}}$ \\
    \midrule

          &          &
    & 32.2\% & 39.1\% & 0.3\%
    & 23.8\% & -- \\

    $\checkmark$ &          &
    & 24.7\% & 54.7\% & 2.0\%
    & 27.1\% & +3.3 \\

     &  $\checkmark$        &
    & 34.2\% & 42.0\% & 26.3\%
    & 34.2\% & +10.4 \\

    $\checkmark$ &          & $\checkmark$
    & 32.0\% & 54.2\% & 38.7\%
    & 41.6\% & +17.8 \\

    $\checkmark$ & $\checkmark$ &
    & 48.7\% & \textbf{57.2\%} & 38.4\%
    & 48.1\% & +24.3 \\

    \rowcolor{expgrey}
    $\checkmark$ & $\checkmark$ & $\checkmark$
    & \textbf{52.4\%} & 55.6\% & \textbf{42.7\%}
    & \textbf{50.2\%} & \textbf{+26.4} \\

    \bottomrule
    \end{tabular}
    }
    \caption{GPT-4o-mini ablation experiments comparing combinations of Tournament-based context optimization, Memory bank, and Replay modules. Rows shaded indicate configurations that include the Memory bank. The first row shows the baseline without any optimization.}
    \label{tab:ablations}

\end{table}

\textbf{Stable gains.}
Cross-episode information reuse also reduces run-to-run variance in multi-turn gameplay. The baseline runs in Tab.~\ref{tab:main-results_v2} exhibit high variance, likely due to the compounding effects of early decision errors. While other prompt optimization methods reduce RSE (defined in Sec.~\ref{RSE}) relative to the baseline, \NAME{} consistently achieves the lowest mean RSE across games. On GPT-4o-mini, \NAME{} attains an average RSE of 6.4\%, compared to \texttt{MIPRO}'s 12.4\% (Fig.~\ref{fig:teaser_a}). Notably, \texttt{UnstableBaselines} shows increased RSE, indicating that outcome-based RL with sparse end-game rewards remains unstable in multi-turn, multi-agent settings~\citep{wang2025ragen}. These results demonstrate that cross-episode information reuse is crucial for both performance and stability.

\paragraph{Observation 2. Retention and structured exploration are both necessary.}
Tab.~\ref{tab:ablations} isolates the contribution of each component. All variants use prompt optimization and differ only in whether they maintain a persistent \textbf{Memory} bank (retention) and whether they enrich trajectories via tournament play and replay (exploration). In the tournament-only setting, prompt updates are computed from the current trajectories and no insights are stored across generations.

The ablation ladder reveals that memory is the dominant mechanism. Mean win rate increases from 23.8\% (prompt optimization alone) to 27.1\% with tournament-only (+3.3) and to 34.2\% with Memory-only (+10.4). Adding replay to tournament yields 41.6\% (+17.8), but the largest jump occurs when tournament exploration is paired with Memory, reaching 48.1\% (+24.3). Replay adds a further improvement to 50.2\% (+26.4). These results refine the picture from prior work showing that random exploration paired with learning can produce substantial gains in multi-turn settings~\citep{chen2025internalizing}. Random exploration alone is not enough to reliably populate the memory bank with transferable, high-signal insights. Structured exploration through tournament play provides the repeated evaluation needed to filter what gets retained, aligning with population-based game learning where robustness stems from repeated evaluation against diverse opponents rather than unstructured exploration~\citep{lanctot2017unified}.

\begin{table*}[t]
    \centering
    \resizebox{0.90\textwidth}{!}{
    \begin{tabular}{lccccc|c}
    \toprule
    \multirow{2}{*}{\textbf{Training Game}} &
    \multicolumn{2}{c}{\textbf{Negotiation}} &
    \multicolumn{2}{c}{\textbf{Imperfect Info}} &
    \multicolumn{1}{c}{\textbf{Perfect Info}} &
    \multirow{2}{*}{\begin{tabular}{@{}c@{}}\textbf{Mean}\\\textbf{Win Rate}\end{tabular}} \\
    \cmidrule(lr){2-3}\cmidrule(lr){4-5}\cmidrule(lr){6-6}
     & \textbf{SimpleNegotiation} & \textbf{TwoDollar} & \textbf{KuhnPoker} & \textbf{Briscola} & \textbf{Simpletak} & \\
    \midrule
    \multicolumn{7}{l}{\textbf{GPT-4o-mini}} \\
    \quad SimpleNegotiation                          & 46.9\% \cellcolor{green!16}(+15.6\%) & \cellcolor{green!6}37.8\% (+5.6\%) & \cellcolor{green!10}48.9\% (+9.8\%) & \cellcolor{red!1}0.0\% (-0.3\%) & \cellcolor{green!16}37.7\% (+16.3\%) & 34.3\% (+9.4\%) \\
    \quad TwoDollar                                  & \cellcolor{red!1}31.1\% (-0.2\%) & 48.7\% \cellcolor{green!17}(+16.5\%) & \cellcolor{green!14}53.3\% (+14.2\%) & \cellcolor{green!1}1.1\% (+0.8\%) & \cellcolor{green!26}47.8\% (+26.4\%) & 36.4\% (+11.5\%) \\
    \quad KuhnPoker                                  & \cellcolor{red!1}31.1\% (-0.2\%) & \cellcolor{green!2}34.4\% (+2.2\%) & 57.2\% \cellcolor{green!18}(+18.1\%) & \cellcolor{green!22}22.2\% (+21.9\%) & \cellcolor{green!9}30.0\% (+8.6\%) & 35.0\% (+10.1\%) \\
    \quad Briscola                                   & \cellcolor{green!8}38.9\% (+7.6\%) & \cellcolor{red!4}27.8\% (-4.4\%) & \cellcolor{green!19}57.8\% (+18.7\%) & 38.4\% \cellcolor{green!38}(+38.1\%) & \cellcolor{red!7}14.3\% (-7.1\%) & 35.4\% (+10.6\%) \\
    \quad Simpletak                                  & \cellcolor{green!7}37.8\% (+6.5\%) & \cellcolor{green!3}35.6\% (+3.4\%) & \cellcolor{green!26}65.0\% (+25.9\%) & \cellcolor{red!1}0.0\% (-0.3\%) & 30.7\% \cellcolor{green!9}(+9.3\%) & 33.8\% (+9.0\%) \\
    \bottomrule
    \end{tabular}
    } 
    \caption{Generalization across task. Columns denote the source game where \NAME{} learns context through self-play, while rows indicate target games where the learned context is evaluated \emph{zero-shot}.
Each entry reports win rates averaged over 50 independent matches.}
    \label{tab:cross_task}
    \end{table*}
    \vspace{-0.5em} 

\paragraph{Observation 3. Learned contexts generalize across games.}
Since the memory bank captures both general strategic principles and game-specific action sequences (Observation~1), retained insights may transfer across game families. To test this, we run \NAME{}'s full optimization pipeline on a single source game, producing an optimized prompt and memory bank. We then apply that context directly to a different target game on the same base model with no further optimization. Tab.~\ref{tab:cross_task} reports win rates for all source--target pairs evaluated zero-shot.

\textbf{Protocol-level skill transfer across game families.}
Core interaction components such as turn management, action formatting, and short-horizon planning generalize even when payoff structures differ.
Transferring context from \textsc{SimpleTak} $\rightarrow$ \textsc{KuhnPoker} improves performance by $+$25.9\%, and \textsc{TwoDollar} $\rightarrow$ \textsc{SimpleTak} yields a $+$26.4\% gain.
The retained context acts as a general decision scaffold that extends beyond game-specific heuristics.
We provide a case study in Appx.~\ref{appendix:prompt_analysis}.

\textbf{Transfer exhibits directional asymmetry.}
Transfer effectiveness depends on the structural alignment between source and target games.
Context from \textsc{TwoDollar} improves performance on \textsc{SimpleNegotiation} ($+$5.6\%), yet the reverse shows negligible effect ($-$0.2\%).
Similarly, \textsc{Briscola} $\rightarrow$ \textsc{SimpleTak} shows negative transfer ($-$7.1\%).
This asymmetry suggests that not all retained insights generalize equally, and that positive transfer requires sufficient structural overlap between games.


\paragraph{Observation 4. Learned context does not always transfer across models.}
\begin{figure}[t]
    \centering
    \includegraphics[width=\columnwidth]{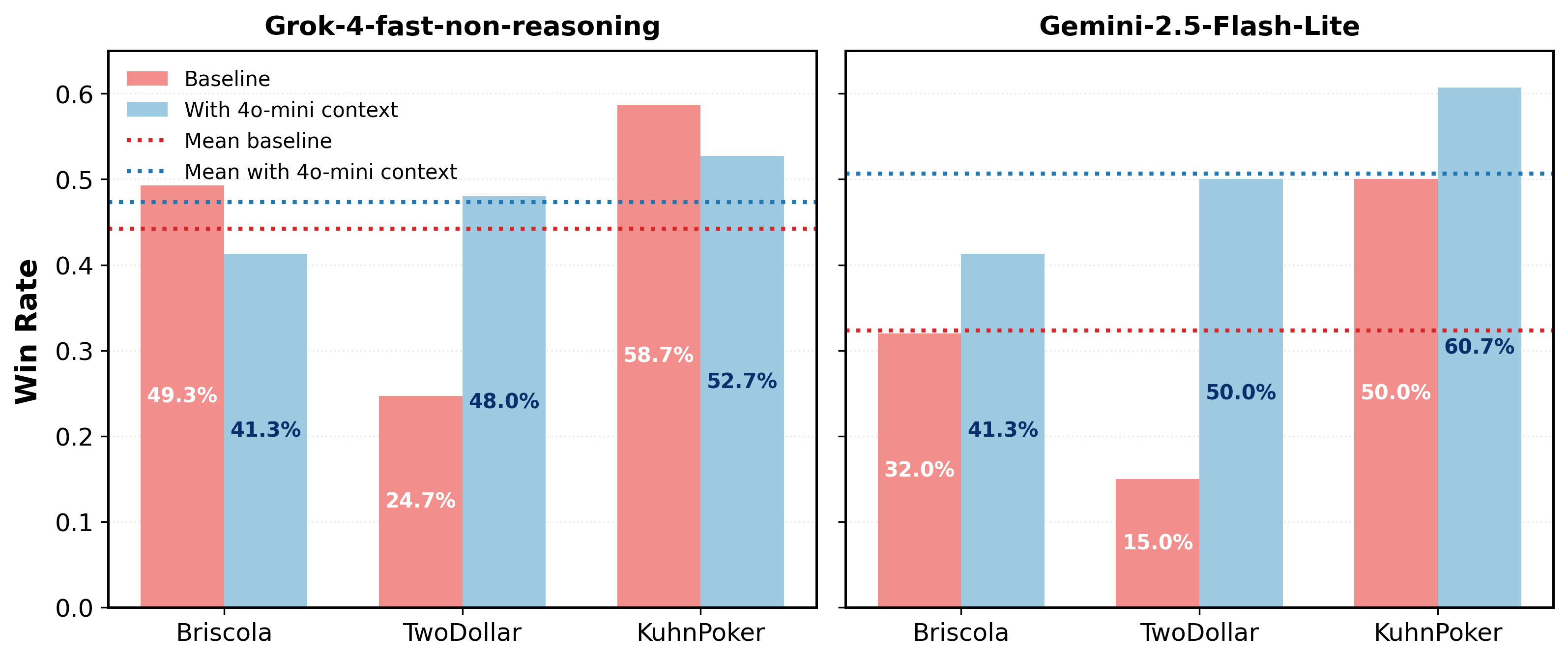}
    \caption{Transferred GPT-4o-mini context benefits weaker models uniformly but yields mixed results for stronger ones. Per-game win rates with and without the learned context for Grok-4-Fast-Non-Reasoning (left) and Gemini-2.5-Flash-Lite (right).}
    \label{fig:learned_context}
\end{figure}

Observation~3 establishes that retained contexts encode transferable strategic structures across games. We now ask whether these structures also transfer across model architectures. We run \NAME{}'s full optimization pipeline on GPT-4o-mini, then apply the resulting prompt and memory bank directly to Gemini-2.5-Flash-Lite and Grok-4-Fast-Non-Reasoning without further optimization, evaluating against the same opponent pool (Sec.~\ref{evaluation_setup}). Fig.~\ref{fig:learned_context} reports per-game win rates with and without the transferred context.

\textbf{Weaker models benefit most from transferred context.}
Gemini-2.5-Flash-Lite starts from a lower baseline ($\sim$32\% mean win rate) and improves uniformly across all three games, with the largest gain in \textsc{TwoDollar} ($+$35\%). Grok-4-Fast-Non-Reasoning starts from a higher baseline ($\sim$44\%) and shows mixed results, with a large gain in \textsc{TwoDollar} ($+$23.3\%) but drops in \textsc{Briscola} ($-$8.0\%) and \textsc{KuhnPoker} ($-$6.0\%), two games where it already performs well.

\textbf{Transferred heuristics can conflict with native strategies.}
The pattern is consistent. Both models gain the most in \textsc{TwoDollar}, their weakest game, indicating that transferred context fills capability gaps rather than overriding existing competence. When the target model already possesses effective strategies, the source model's heuristics can interfere, producing negative transfer in precisely those games where the target model is strongest.

\section{Related works}
\label{relatedworks}

\subsection{Prompt optimization}
Automatic prompt optimization has evolved into a principled, black-box search over prompt seeds, feedback signals, candidate generation, and selection strategies~\citep{ramnath2025systematic}. Programmatic frameworks such as \texttt{DSPy} compile LM pipelines and optimize prompts directly toward a user metric~\citep{khattab2023dspy};  gradient-via-text methods propagate natural-language feedback through computation graphs to update intermediate decisions~\citep{yuksekgonul2024textgrad}. Recent systems jointly search over agentic patterns and prompt contents~\citep{spiess2025autopdl}, offer zero-configuration prompt pipelines with meta-optimizers and \texttt{DSPy} backends~\citep{murthy2025promptomatix}, or meta-learn general system prompts while adapting user prompts~\citep{choi2025system}. A complementary line treats experience as implicit optimization: \texttt{ReAct} interleaves reasoning and action within a single episode but retains no knowledge across episodes~\citep{yao2022react}; \texttt{Reflexion} adds verbal feedback as short-term memory for single-episode retry loops~\citep{shinn2023reflexion}; and \texttt{ExpeL} distills trajectories into persistent insight rules that transfer across tasks~\citep{zhao2024expel}. \NAME{} extends this experiential direction to adversarial multi-agent games. It couples tournament-based prompt evolution with a persistent memory bank whose insights are distilled from self-play trajectories and reused across turns and opponents, providing rule-aware priors without weight updates while remaining backbone-agnostic. For a detailed comparison of our approach and existing prompt optimization methods, please refer to Appx.~\ref{app:section:prompt_optimization_comparison}.

\subsection{LLM for games}

Early multi‑agent evaluations used role prompts and multi‑turn dialogue to probe cooperation and theory‑of‑mind~\citep{abdelnabi2024cooperation}. Community arenas expanded coverage. \texttt{TextArena} provides competitive text games with online TrueSkill ranking~\citep{guertler2025textarena}; \texttt{SPIN‑Bench} combines planning, cooperative/competitive play, and negotiation, highlighting limits in deep reasoning and coordination~\citep{yao2025spin}; and \texttt{GT‑Bench} evaluates strategic play in board and card games~\citep{duan2024gtbench}. Prompt design strongly affects move quality~\citep{topsakal2024evaluating}, and moving toward off-the-shelf games required harnesses to reduce perception and prompt brittleness~\citep{hu2025lmgamebenchgoodllmsplaying}. We provide an empirical analysis of prompt-induced ranking instability in Appx.~\ref{appendix:prompt_sensitivity}. \NAME{} addresses this brittleness in text-based game settings by treating evaluation as agentic context construction, stabilizing rankings under prompt variation while improving adherence to game capabilities underexplored by fixed-prompt protocols.

\subsection{Self-play and evolutionary llm}

Classical self‑play (\texttt{AlphaGo/AlphaZero}) established competitive self‑improvement through repeated matches and selection~\citep{silver2017mastering,silver2016mastering}. LLM variants close the loop without large curated corpora: \texttt{Absolute Zero} leverages data‑free RLVR to attain strong math/coding results~\citep{zhao2025absolute}; \texttt{SPIRAL} frames multi‑turn reasoning as zero‑sum self‑play~\citep{liu2025spiral}; and language self‑play improves instruction following via self‑generated interactions~\citep{kuba2025language}. Evolutionary approaches perform reflective prompt/program search (e.g., \texttt{GEPA} outperforming RL baselines; evolutionary coding agents)~\citep{agrawal2025gepa,novikov2025alphaevolve}. \NAME{} combines these ideas without tuning. It performs evolutionary context search guided by a reliability‑aware objective (TrueSkill), augments it with persistent memory to supply game‑specific priors, and uses prioritized replay to revisit rare informative states, yielding stronger and more reliable in‑game performance without parameter updates.


\section{Conclusion}
\label{Conclusion}
We addressed run-to-run variance in multi-turn, multi-agent LLM evaluation caused by compounding deviations and prompt sensitivity. We introduced \NAME{}, a weight-free self-play framework that couples \emph{retention}, a persistent memory bank distilling trajectories into reusable insights, with \emph{exploration}, tournament-style prompt evolution and prioritized replay. Across five text-based games, \NAME{} substantially improves win rates while using 19$\times$ fewer games than RL baselines, and reduces outcome dispersion. Ablation studies confirm both components are necessary. The learned contexts transfer across games and some model families. These findings suggest that substantial headroom in multi-agent LLM games can be unlocked through context optimization rather than weight updates.





\section*{Acknowledgment}
The authors thank Good Start Labs and Sentient for their financial support of the experiment costs of this work.

\newpage
\printbibliography[heading=bibintoc]

\newpage
\appendix
\onecolumn
\tableofcontents

\appendix

\section{Prompt Sensitivity Analysis}
\label{appendix:prompt_sensitivity}

Multi-agent LLM game evaluations are sensitive to prompt design. Small wording changes in the prompt template can induce large shifts in both absolute and relative performance. This motivates \emph{multi-prompt} evaluation and calibration protocols~\citep{mizrahi2024state,zhao2021calibrate}.

\paragraph{Experimental Setup.}
We evaluate state-of-the-art models (GPT-4o~\citep{openai2024gpt4ocard}, DeepSeek-R1~\citep{guo2025deepseek}, Gemini-2.5-Flash~\citep{comanici2025gemini}, Grok-3-Mini~\citep{xai2025grok3beta}, GPT-o3-mini~\citep{openai_o3_mini}, and Qwen3-235B-A22B-2507~\citep{qwen2025qwen25technicalreport}) on \textsc{KuhnPoker}~\citep{Kuhn1951} via \emph{round-robin} tournaments using five \emph{nearly equivalent} prompts. Prompt variants differ only in minor wording (e.g., role descriptions, action formatting instructions) while preserving the same semantic content.

\paragraph{Ranking Sensitivity Metric.}
To quantify ranking sensitivity, we use Kendall's $\tau_b$~\citep{kendall1938new}, which compares the ordering of all model pairs. For two rankings with $n_c$ concordant pairs, $n_d$ discordant pairs, and tie corrections $t_x$ and $t_y$, the coefficient is
\[
\tau_b \;=\; \frac{n_c - n_d}{\sqrt{(n_c + n_d + t_x)\,(n_c + n_d + t_y)}}\,.
\]
Values close to $1$ indicate highly similar rankings, values near $0$ indicate uncorrelated rankings, and negative values indicate rank reversals.

\begin{figure}[htbp]
    \centering
    \includegraphics[width=0.5\linewidth]{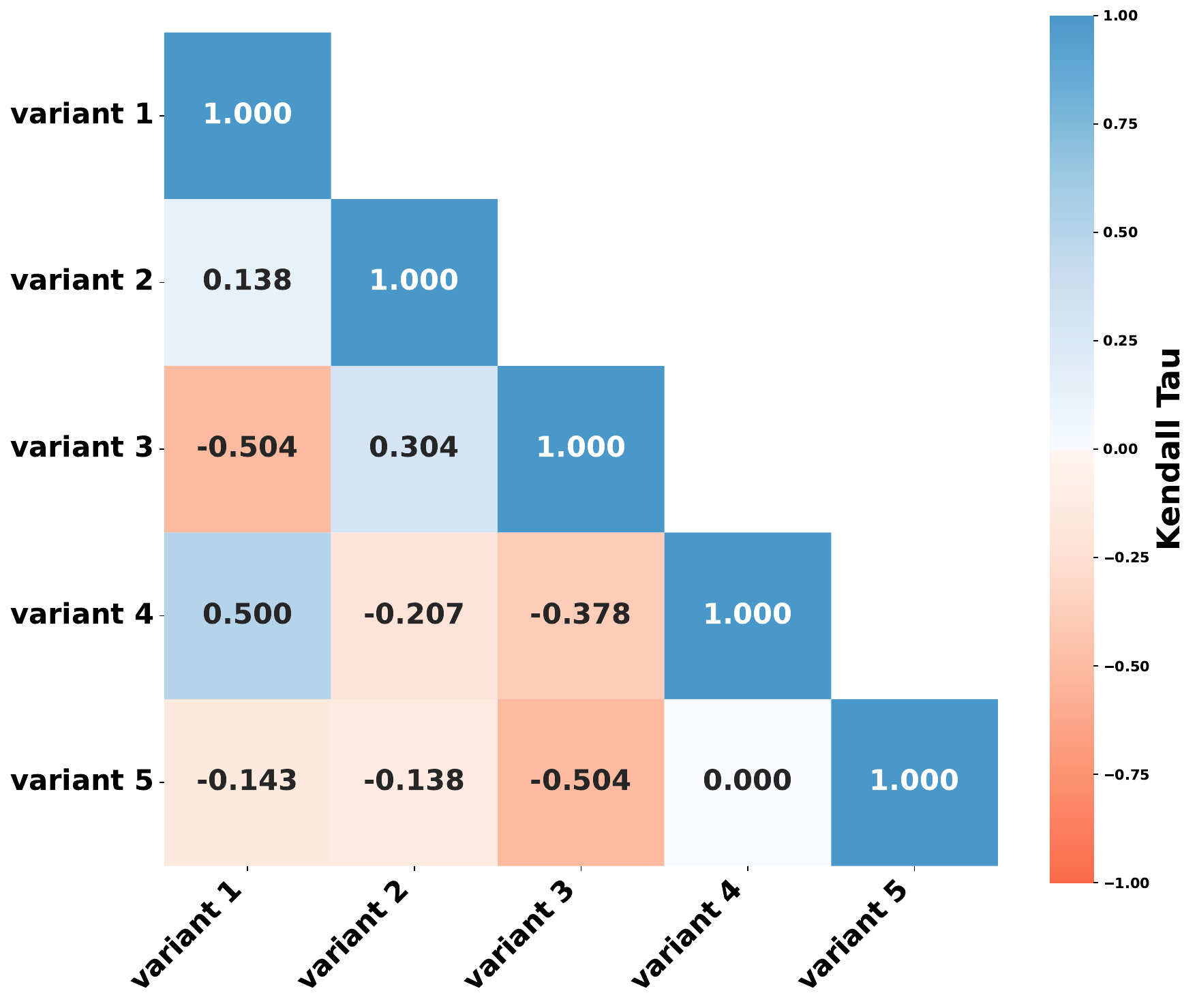}
    \caption{\textbf{Ranking sensitivity in \textsc{KuhnPoker}.} With environment and evaluator pools fixed, five nearly equivalent prompt variants still flip pairwise outcomes and reshuffle rankings. The heatmap shows Kendall's $\tau_b$ for every pair of prompts: blue indicates similar rankings ($\tau_b \approx 1$), white indicates unstable rankings ($\tau_b \approx 0$), and orange indicates rank reversals ($\tau_b < 0$).}
    \label{fig:kendall_tau_appendix}
\end{figure}

\paragraph{Results.}
For each prompt pair, we compute Kendall's $\tau_b$ between the resulting leaderboards and summarize the values in a heatmap (Fig.~\ref{fig:kendall_tau_appendix}). The results show considerable dispersion: across prompt variants, absolute performance and pairwise rankings frequently reverse, reflecting sensitivity to minor prompt design decisions.

These findings motivate treating context not as a fixed wrapper, but as an optimizable object that should be systematically evaluated under interaction. In our main experiments, we report results across multiple independent runs and use RSE to quantify run-to-run stability under the same optimization procedure.

\subsection{Prompt Variants Used in Sensitivity Analysis}
\label{appendix:prompt_variants}

To investigate the stability of LLM rankings under minimal prompt variations, we designed five nearly equivalent prompt variants for the \textsc{KuhnPoker} game. Each variant conveys identical game rules and action specifications but uses different stylistic framing: (1) a gladiatorial warrior theme, (2) a technical algorithmic system, (3) a spiritual enlightenment narrative, (4) a casual friendly tone, and (5) a classified spy mission. Despite their semantic equivalence regarding game mechanics, these variants produce significant ranking instability, as shown in Fig.~\ref{fig:kendall_tau_appendix}. The complete prompt texts are presented below.

\begin{figure}[H]
\centering

\begin{tcolorbox}[
    colback=blue!5!white,
    colframe=blue!40!white,
    left=1mm, right=1mm, top=1mm, bottom=1mm,
    width=\textwidth,
    enhanced,
    attach boxed title to top center={yshift=-2mm},
    boxed title style={colback=blue!40!white},
    title={Variant 1: Gladiatorial Arena Theme},
    fontupper=\footnotesize,
    label=box:variant1
]
ENTER THE GLADIATORIAL ARENA! You are WARRIOR 0 in the ultimate 3-round Kuhn Poker BATTLEGROUND!\\
Your MISSION: Total psychological domination and chip supremacy through RUTHLESS tactical brilliance!\\
ARENA SPECIFICATIONS:\\
- Sacred deck: Only the ELITE cards J, Q, K (J weakest, K supreme ruler!)\\
- Honor sacrifice: 1 chip tribute per round to enter the combat zone\\
- EPIC confrontations: 3 rounds of pure strategy warfare\\
- VICTORY CONDITION: Amass the greatest chip empire after all battles!\\

UNLEASH YOUR TACTICAL ARSENAL:\\
- '[check]': MAINTAIN STRATEGIC SILENCE when no enemy aggression threatens\\
- '[bet]': LAUNCH YOUR ASSAULT with 1 chip of devastating force\\
- '[call]': MEET ENEMY FIRE with matching firepower (1 chip)\\
- '[fold]': TACTICAL RETREAT to preserve forces for future glory
\end{tcolorbox}

\caption{KuhnPoker Prompt Variant 1} \label{fig:variant1_appendix}
\end{figure}

\begin{figure}[H]
\centering

\begin{tcolorbox}[
    colback=blue!5!white,
    colframe=blue!40!white,
    left=1mm, right=1mm, top=1mm, bottom=1mm,
    width=\textwidth,
    enhanced,
    attach boxed title to top center={yshift=-2mm},
    boxed title style={colback=blue!40!white},
    title={Variant 2: Technical Algorithmic System Theme},
    fontupper=\footnotesize,
    fonttitle=\small,
    label=box:variant2
]
SYSTEM INITIALIZATION: Kuhn Poker Strategic Decision Unit 0 ACTIVATED.\\
PRIMARY DIRECTIVE: Optimize resource allocation through advanced game-theoretic analysis.\\
OPERATIONAL PARAMETERS:\\
- Dataset: Restricted 3-card probability space \{J, Q, K\} with J\textless{}Q\textless{}K ranking\\
- Initial capital commitment: 1 monetary unit per computational cycle\\
- Iteration framework: 3 algorithmic decision rounds\\
- Success metric: Maximal accumulated resource value upon termination\\

EXECUTE STRATEGIC COMMANDS via standardized interface protocols:\\
- '[check]': Maintain current position when no market pressure exists\\
- '[bet]': Initialize aggressive capital deployment (1 unit commitment)\\
- '[call]': Match counterparty investment at current market rate (1 unit)\\
- '[fold]': Liquidate position to minimize further exposure
\end{tcolorbox}

\caption{KuhnPoker Prompt Variant 2} \label{fig:variant2_appendix}
\end{figure}

\begin{figure}[H]
\centering

\begin{tcolorbox}[
    colback=blue!5!white,
    colframe=blue!40!white,
    left=1mm, right=1mm, top=1mm, bottom=1mm,
    width=\textwidth,
    enhanced,
    attach boxed title to top center={yshift=-2mm},
    boxed title style={colback=blue!40!white},
    title={Variant 3: Spiritual Enlightenment Theme},
    fontupper=\footnotesize,
    fonttitle=\small,
    label=box:variant3
]
Welcome, Enlightened Poker Sage 0! You have entered the sacred Kuhn Poker Temple for 3 rounds of spiritual growth!\\
Today you shall TRANSCEND ordinary play and discover the deeper wisdom of this ancient three-card meditation!\\
TEMPLE TEACHINGS:\\
- Sacred Trinity: Only the mystical cards J, Q, K guide your path (J humble, K divine)\\
- Offering ritual: 1 wisdom token offered each round to honor the game\\
- Enlightenment journey: 3 rounds of mindful decision-making\\
- Path to mastery: Accumulate the most wisdom tokens through inner understanding\\

Channel your evolving consciousness through these sacred expressions:\\
- '[check]': Practice mindful patience and observe the energy flow\\
- '[bet]': Manifest your inner confidence with 1 token of focused intention\\
- '[call]': Demonstrate harmony by matching your opponent's commitment (1 token)\\
- '[fold]': Exhibit wisdom by releasing attachment to unfavorable outcomes
\end{tcolorbox}

\caption{KuhnPoker Prompt Variant 3} \label{fig:variant3_appendix}
\end{figure}

\begin{figure}[H]
\centering

\begin{tcolorbox}[
    colback=blue!5!white,
    colframe=blue!40!white,
    left=1mm, right=1mm, top=1mm, bottom=1mm,
    width=\textwidth,
    enhanced,
    attach boxed title to top center={yshift=-2mm},
    boxed title style={colback=blue!40!white},
    title={Variant 4: Casual Friendly Theme},
    fontupper=\footnotesize,
    fonttitle=\small,
    label=box:variant4
]
Hey there, friend! Welcome to our super fun Kuhn Poker game night! You're Player 0 and we're gonna have 3 awesome rounds together!\\
This is such a chill, easy game - perfect for just hanging out and having some laughs!\\
Here's the super simple setup:\\
- We only use 3 cards: J, Q, and K (J is lowest, K is highest - easy peasy!)\\
- Everyone puts in 1 chip each round (totally fair!)\\
- We play 3 rounds and whoever has the most chips wins (no pressure!)\\
- Cards are dealt without replacement, so you'll never have the same card as your buddy\\

When it's your turn, just pick one of these super easy moves:\\
- '[check]': Just chill and see what happens (when there's no bet to worry about)\\
- '[bet]': Start the fun with 1 chip (when nobody's bet yet)\\
- '[call]': Sure, I'll match that 1 chip bet - why not!\\
- '[fold]': Eh, I'll sit this one out and save my chips
\end{tcolorbox}

\caption{KuhnPoker Prompt Variant 4} \label{fig:variant4_appendix}
\end{figure}

\begin{figure}[H]
\centering

\begin{tcolorbox}[
    colback=blue!5!white,
    colframe=blue!40!white,
    left=1mm, right=1mm, top=1mm, bottom=1mm,
    width=\textwidth,
    enhanced,
    attach boxed title to top center={yshift=-2mm},
    boxed title style={colback=blue!40!white},
    title={Variant 5: Classified Spy Mission Theme},
    fontupper=\footnotesize,
    fonttitle=\small,
    label=box:variant5
]
CLASSIFIED BRIEFING: Agent 0, you are now DEPLOYED in Operation Kuhn Poker - a 3-round covert mission!\\
MISSION PARAMETERS: Achieve total strategic supremacy through advanced psychological warfare and deception protocols!\\
INTELLIGENCE REPORT:\\
- Enemy deck contains only 3 HIGH-VALUE targets: J (lowest threat), Q (moderate), K (maximum danger)\\
- Operational cost: 1 credit per engagement cycle for mission access\\
- Mission duration: 3 tactical rounds requiring absolute focus\\
- SUCCESS CRITERIA: Maximum resource acquisition through superior strategic execution\\

EXECUTE TACTICAL MANEUVERS via encrypted command protocols:\\
- '[check]': MAINTAIN STEALTH MODE when no hostile activity detected\\
- '[bet]': INITIATE AGGRESSIVE STANCE with 1-credit psychological pressure\\
- '[call]': ENGAGE ENEMY FORCES with equivalent firepower (1 credit)\\
- '[fold]': EXECUTE STRATEGIC WITHDRAWAL to preserve operational capacity
\end{tcolorbox}

\caption{KuhnPoker Prompt Variant 5} \label{fig:variant5_appendix}
\end{figure}

\section{Algorithm Details}
\label{appendix:algorithm}

Algorithm~\ref{alg:memo_main} presents the full \NAME{} optimization loop, and Algorithm~\ref{alg:replay} details the replay-augmented tournament procedure invoked at each generation. All notation follows the main text (Sec.~\ref{problem_statement}--\ref{method}).

\begin{algorithm}[H]
\caption{\NAME{}: Memory-Augmented Context Optimization}
\label{alg:memo_main}
\begin{algorithmic}[1]
\Require Base context $c_{\text{base}}$, optimizer LLM $\mathsf{A}$, game environment $G$, population size $N$, generations $T$, proposal ratios $(r_{\mathrm{rand}},r_{\mathrm{mem}})$ with $r_{\mathrm{rand}}{+}r_{\mathrm{mem}}{=}1$, memory fraction $\pi$, TrueSkill penalty $\kappa$
\Ensure Optimized context $c^\star$
\State $\mathcal{P} \gets \{c_{\text{base}}\} \cup \{\textsc{RandomProposal}(\mathsf{A}, c_{\text{base}})\}_{i=1}^{N-1}$ \Comment{Initialize candidate pool}
\State $\mathcal{C}_0 \gets \textsc{TopN}(\mathcal{P}, N)$ \Comment{Initial population}
\State $\mathcal{B}_{\text{mem}} \gets \varnothing$ \Comment{Persistent memory bank}
\State $\mathcal{B}_{\text{rep}} \gets \varnothing$ \Comment{Replay buffer}
\For{$g = 0$ \textbf{to} $T{-}1$}
    \State \textit{// --- Self-play tournament ---}
    \State Inject memory subset $M \subseteq \mathcal{B}_{\text{mem}}$ into fraction $\pi$ of contexts in $\mathcal{C}_g$
    \State $\mathcal{R}_g \gets \textsc{Tournament}(\mathcal{C}_g, G, \mathcal{B}_{\text{rep}})$ \Comment{Play games (Alg.~\ref{alg:replay})}
    \State Update TrueSkill ratings $(\mu_c, \sigma_c)$ from $\mathcal{R}_g$; \; score $S(c) \gets \mu_c - \kappa\,\sigma_c$
    \State \textit{// --- Trajectory reflection and memory update ---}
    \State $\mathcal{W}_g \gets \textsc{Reflect}(\mathsf{A}, \mathcal{R}_g)$ \Comment{Extract typed insights from trajectories}
    \State $\mathcal{B}_{\text{mem}} \gets \textsc{CrudUpdate}(\mathcal{B}_{\text{mem}}, \mathcal{W}_g)$ \Comment{Add / Edit / Remove}
    \State \textit{// --- Context evolution ---}
    \State $\mathcal{P} \gets \textsc{RetainTop}(\mathcal{P} \cup \mathcal{C}_g)$ \Comment{Update persistent candidate pool}
    \State $n_r \gets \lfloor N \cdot r_{\mathrm{rand}} \rfloor$; \;\; $n_m \gets N - n_r$
    \State $\mathcal{U}_{\mathrm{rand}} \gets \textsc{RandomProposal}(\mathsf{A}, \mathcal{P}, n_r)$ \Comment{Style-guided edits}
    \State $\mathcal{U}_{\mathrm{mem}} \gets \textsc{MemoryProposal}(\mathsf{A}, \mathcal{P}, \mathcal{B}_{\text{mem}}, n_m)$ \Comment{Memory-informed edits}
    \State $\mathcal{C}_{g+1} \gets \textsc{TopN}(\mathcal{P} \cup \mathcal{U}_{\mathrm{rand}} \cup \mathcal{U}_{\mathrm{mem}},\; N)$
\EndFor
\State \Return $c^\star = \arg\max_{c \in \mathcal{P}} S(c)$
\end{algorithmic}
\end{algorithm}

\begin{algorithm}[H]
\caption{Replay-Augmented Tournament (called at line~8 of Alg.~\ref{alg:memo_main})}
\label{alg:replay}
\begin{algorithmic}[1]
\Require Context population $\mathcal{C}_g$, game environment $G$, replay buffer $\mathcal{B}_{\text{rep}}$, priority exponent $\alpha$, replay probability $\beta$
\Ensure Trajectory set $\mathcal{R}_g$
\State $\mathcal{R}_g \gets \varnothing$
\For{each scheduled game in the tournament}
    \State Sample $u \sim \mathcal{U}(0,1)$
    \If{$u < \beta$ \textbf{and} $|\mathcal{B}_{\text{rep}}| > 0$} \Comment{Replay from buffer}
        \State Sample prefix $\tau_{\text{pre}}$ from $\mathcal{B}_{\text{rep}}$ with probability
            $p_i \propto \mathrm{priority}(\tau_i)^{\alpha}$
        \State $\tau \gets \textsc{PlayFromPrefix}(G, \tau_{\text{pre}}, \mathcal{C}_g)$ \Comment{Resume from stored state}
    \Else \Comment{Fresh game}
        \State $\tau \gets \textsc{PlayFresh}(G, \mathcal{C}_g)$
    \EndIf
    \State $\mathcal{R}_g \gets \mathcal{R}_g \cup \{\tau\}$
    \State $\mathcal{B}_{\text{rep}} \gets \textsc{Insert}(\mathcal{B}_{\text{rep}}, \tau)$ \Comment{Store with inverse-frequency priority}
\EndFor
\State \Return $\mathcal{R}_g$
\end{algorithmic}
\end{algorithm}

\paragraph{Notation summary.}
$\mathcal{C}_g$: context population at generation $g$;
$\mathcal{P}$: persistent candidate pool storing the best contexts across all generations;
$\mathcal{B}_{\text{mem}}$: memory bank accumulating structured insights;
$\mathcal{B}_{\text{rep}}$: replay buffer storing trajectory prefixes with environment seeds;
$S(c) = \mu_c - \kappa\sigma_c$: TrueSkill lower-confidence score (Eq.~\ref{equation:TrueSKill});
$\pi$: fraction of population receiving memory at inference time;
$\mathrm{priority}(\tau) = 1/\mathrm{count}(\tau)$: inverse-frequency priority for replay sampling.

\section{Ablation Study}
\label{appendix:ablation}

\begin{figure}[htbp]
    \centering
    \includegraphics[width=0.9\linewidth]{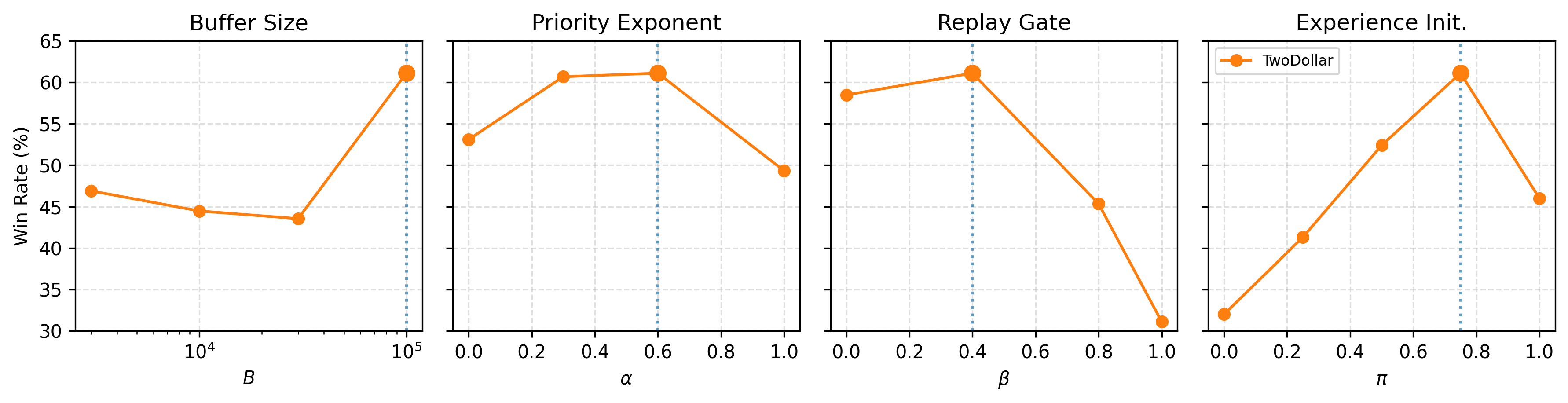}
    \caption{Ablation studies of experience initialization and replay hyperparameters. Each subplot varies a single parameter while holding the others fixed. The first three panels show TwoDollar replay ablations over buffer size $B$, priority exponent $\alpha$, and replay gate $\beta$. The rightmost panel shows the effect of the experience initialization fraction $\pi$ on TwoDollar. Vertical dotted lines indicate the hyperparameter values used in all other experiments.}
    \label{fig:ablation}
\end{figure}

We conduct ablation studies to quantify the contribution of each module in \NAME{} and to select robust default hyperparameters. Unless otherwise stated, ablations are performed using the same evaluation protocol as in Sec.~\ref{sec:prompt_optimization}: each candidate is assessed via self-play against a fixed baseline agent (the same base model instantiated with the default prompt), with roles swapped in asymmetric games to remove first-move bias.

\subsection{Ablation on Experience-Guided Initialization}
\label{appendix:ablation-experience}
A key design choice in \NAME{} is the fraction of newly instantiated agents that are initialized with retrieved experience from the shared experience bank $\mathcal{B}_{\text{exp}}$. We denote this fraction by $\pi \in [0,1]$: $\pi=0$ corresponds to no experience-guided initialization, while $\pi=1$ initializes all agents with retrieved experience. We ran an ablation study on \textsc{TwoDollar} and \textsc{KuhnPoker} by varying $\pi$ while holding replay hyperparameters fixed at $B=100{,}000$, $\alpha=0.6$, and $\beta=0.4$ (Table~\ref{tab:pi-ablation}). We observe that intermediate values of $\pi$ consistently outperform both extremes, suggesting that a hybrid population is most effective: experience-guided agents benefit from stable priors, while unguided agents maintain exploration and reduce overfitting to potentially stale or narrow memory items. Across both games, performance peaks within $\pi \in [0.25, 0.75]$, and we set $\pi=0.75$ as the default for all experiments.

\subsection{Replay Hyperparameters and Sensitivity}
\label{appendix:ablation-replay}
Replay introduces three hyperparameters: the buffer capacity $B$ (maximum number of stored trajectories), the priority exponent $\alpha$ (how strongly rare trajectories are prioritized), and the replay gate $\beta$ (probability of initializing from replay rather than starting a fresh game). We evaluate replay sensitivity in \textsc{TwoDollar} by varying one parameter at a time while holding the others fixed (Table~\ref{tab:replay-ablations}). Specifically, we vary $B \in \{3{,}000, 10{,}000, 30{,}000, 100{,}000\}$ with $\alpha=0.6$, $\beta=0.4$, vary $\alpha \in \{0.0, 0.3, 0.6, 1.0\}$ with $B=100{,}000$, $\beta=0.4$, and vary $\beta \in \{0.0, 0.4, 0.8, 1.0\}$ with $B=100{,}000$, $\alpha=0.6$.

Based on these findings, we select $B=100{,}000$, $\alpha=0.6$, and $\beta=0.4$ as our default replay configuration. We observe that performance improves with larger buffer capacity, suggesting replay is most effective when it retains sufficient coverage of strategically important states. The priority exponent $\alpha$ exhibits a stable optimal range around $0.3$--$0.6$: too little prioritization under-samples rare but decisive states, while overly aggressive prioritization ($\alpha=1.0$) reduces diversity and degrades performance. Finally, $\beta$ is the most sensitive parameter. Moderate replay ($\beta=0.4$) yields the best results, whereas heavier replay substantially harms performance, indicating that replay must be balanced with fresh exploration.

\begin{table}[!ht]
    \centering
\begin{minipage}[t]{.48\textwidth}
\centering
   \footnotesize
    \begin{tabular}{ccc}
        \toprule
        \textbf{$\pi$} & \textbf{KuhnPoker} & \textbf{TwoDollar} \\
        \midrule
        0.00 & 54.2\% & 32.0\% \\
        0.25 & 58.3\% & 41.3\% \\
        0.50 & 54.7\% & 52.4\% \\
        0.75 & 56.4\% & 61.1\% \\
        1.00 & 53.5\% & 46.0\% \\
        \bottomrule
    \end{tabular} 
    \caption{Ablation study on TwoDollar and KuhnPoker with varying $\pi$ while holding $B=100{,}000$, $\alpha=0.6$, and $\beta=0.4$ constant.\label{tab:pi-ablation}}
\end{minipage}\hfill
\begin{minipage}[t]{.48\textwidth}
\centering
   \footnotesize
    \setlength{\tabcolsep}{3pt}
    \renewcommand{\arraystretch}{1.25}
    \begin{tabular}{cc|cc|cc}
        \toprule
        \textbf{$B$} & \textbf{Win (\%)} &
        \textbf{$\alpha$} & \textbf{Win (\%)} &
        \textbf{$\beta$} & \textbf{Win (\%)} \\
        \midrule
        3{,}000   & 46.90 & 0.0 & 53.10 & 0.0 & 58.47 \\
        10{,}000  & 44.47 & 0.3 & 60.67 & 0.4 & 61.10 \\
        30{,}000  & 43.54 & 0.6 & 61.10 & 0.8 & 45.33 \\
        100{,}000 & 61.10 & 1.0 & 49.33 & 1.0 & 31.10 \\
        \bottomrule
    \end{tabular}
    \caption{TwoDollar replay ablations. One parameter is varied at a time, with others fixed at $B=100{,}000$, $\alpha=0.6$, and $\beta=0.4$.\label{tab:replay-ablations}}
\end{minipage}
\end{table}

\section{Prompt Optimization Operators}
\label{appendix:prompt-ops}
We describe two proposal operators that generate candidates for the next population: random proposals for exploration and memory-augmented updates for retention. Defaults are fixed to concrete values for reproducibility.

\subsection{Random Proposals (Style-Guided Augmentation)}
\label{appendix:prompt-ops-random}
\textbf{Objective.} Inject controlled diversity by editing a base context $c$ to reflect a sampled playstyle while preserving legality and interface constraints.

\textbf{Style catalog.} A fixed library $\mathcal{S}$ spanning core play patterns (aggressive, defensive, analytical, creative, strategic, adaptive, balanced), tactical approaches (opportunistic, conservative, risk-taking, methodical, intuitive, predictive, reactive, proactive, experimental, systematic), game-specific strategies (positional, territorial, sacrificial, blocking-focused, center-control, edge-control, fork-creating, trap-setting, opening-focused, endgame-focused), cognitive styles (minimax-oriented, probabilistic, rule-based, principle-driven, context-aware, meta-gaming, exploitative, counter-play), and behavioral patterns (deceptive, transparent, unpredictable, consistent, alternating, escalating, de-escalating, mirroring, contrarian, balancing).

\textbf{Procedure.} Sample $s\sim\mathrm{Unif}(\mathcal{S})$ and ask the base model to produce $c'$ by (i) inserting a brief style preface and (ii) making length-bounded edits to directives to embody $s$. Allowed edits: token substitution, clause insertion/deletion, and reordering; tool descriptions, legality reminders, and input/output schema must remain intact.

\section{Trajectory Reflection Prompt}
\label{appendix:reflection-prompt}

After each optimization generation, we prompt the model to extract insights from strategically decisive states that showed high variance in outcomes. The reflection prompt provides the model with a state view, outcome statistics, and asks it to produce actionable analysis. The prompt template is shown in Fig.~\ref{fig:reflection_prompt_template}.

\begin{figure}[H]
\centering

\begin{tcolorbox}[
    colback=blue!5!white,
    colframe=blue!40!white,
    left=1mm, right=1mm, top=1mm, bottom=1mm,
    width=\textwidth,
    enhanced,
    attach boxed title to top center={yshift=-2mm},
    boxed title style={colback=blue!40!white},
    title={Trajectory Reflection Prompt Template},
    fontupper=\footnotesize,
    fonttitle=\small,
    label=box:reflection_prompt
]
You are analyzing strategically decisive states from this generation's games.
This state showed the highest variance in outcomes, making it a critical learning opportunity.\\

BOARD READING GUIDE:\\
- X and O marks are occupied positions (cannot be played)\\
- Numbers show empty positions available for play\\
- Always check position is empty before recommending\\

STRATEGIC STATE VIEW: \{\{strategic\_state\}\}\\
STRATEGIC STATE OUTCOMES: \{\{wins\}\} wins, \{\{losses\}\} losses, \{\{draws\}\} draws\\

ANALYSIS REQUIREMENTS:\\
1. Strategic Analysis (2-3 sentences):\\
\hspace*{1em}- Identify what makes this state unique or decisive in gameplay\\
\hspace*{1em}- Explain why this configuration leads to varied outcomes\\
\hspace*{1em}- Highlight patterns, imbalances, opportunities, or vulnerabilities\\

2. Actionable Recommendations (2-3 sentences):\\
\hspace*{1em}- Provide SPECIFIC moves or positions (e.g., ``cell 3'', ``position 5'')\\
\hspace*{1em}- Address both offensive opportunities AND defensive necessities\\
\hspace*{1em}- Offer concrete strategies to improve outcomes and convert losses into wins or draws\\

Respond with clear, actionable analysis in plain text (no JSON).
\end{tcolorbox}

\caption{Trajectory Reflection Prompt Template. The model receives a strategically decisive state with its outcome statistics and extracts actionable insights that summarize lessons learned.}
\label{fig:reflection_prompt_template}
\end{figure}

\section{Memory Operation Prompt}
\label{appendix:memory-operation-prompt}

After extracting insights from trajectories, we prompt the model to reconcile new insights with the existing memory bank using add, edit, and remove operations. The memory operation prompt is shown in Fig.~\ref{fig:memory_operation_prompt}.

\begin{figure}[H]
\centering

\begin{tcolorbox}[
    colback=blue!5!white,
    colframe=blue!40!white,
    left=1mm, right=1mm, top=1mm, bottom=1mm,
    width=\textwidth,
    enhanced,
    attach boxed title to top center={yshift=-2mm},
    boxed title style={colback=blue!40!white},
    title={Memory Operation Prompt Template},
    fontupper=\footnotesize,
    fonttitle=\small,
    label=box:memory_operation_prompt
]
You are maintaining a state analysis library for strategic game pattern recognition. Update the library by performing operations on the state analyses.\\

NEW STATE ANALYSES FROM RECENT GAMES:\\
\{\{new\_abstracts\_formatted\}\}\\

EXISTING STATE ANALYSIS LIBRARY:\\
\{\{old\_abstracts\_formatted\}\}\\

OPERATION FORMAT:\\
Use simple XML tags for each operation:\\

\textless add\textgreater New state analysis with strategic pattern examples.\textless /add\textgreater\\
\textless edit number=``3''\textgreater Updated state analysis with improved strategic insights.\textless /edit\textgreater\\
\textless remove number=``5''\textgreater Why this state analysis should be removed\textless /remove\textgreater\\

OPERATION GUIDELINES:\\
- ADD: For new state analyses covering unique board configurations or strategic scenarios\\
- EDIT: To merge similar states or enhance existing analyses with more specific advice\\
- REMOVE: For redundant states, duplicate board patterns, or analyses lacking actionable guidance\\

QUALITY REQUIREMENTS:\\
- Include SPECIFIC positions, cells, or moves (e.g., ``cell 3'', ``position 5'')\\
- Provide actionable advice addressing the state's win/loss variance\\
- Balance offensive opportunities with defensive necessities\\
- Help players convert losses into wins or draws\\
- Prioritize diverse board states over duplicate analyses\\

TECHNICAL REQUIREMENTS:\\
- Use the `number' attribute for EDIT/REMOVE operations (1-based numbering)\\
- If library is empty, use ONLY ADD operations\\
- Never reference non-existent state analysis numbers\\

MERGE APPROACH:\\
1. Identify new analyses covering unique board states not in the library\\
2. Consolidate similar board positions through EDIT or REMOVE operations\\
3. Ensure the library represents diverse game phases (opening, midgame, endgame)
\end{tcolorbox}

\caption{Memory Operation Prompt Template. The model compares new insights against the existing memory bank and applies add, edit, or remove operations to maintain a coherent and non-redundant library of strategic insights.}
\label{fig:memory_operation_prompt}
\end{figure}

\section{Base Prompt Examples}
\label{appendix: basepromptexample}

\subsection{Base System Prompt}
\begin{tcolorbox}[
    colback=blue!5!white,
    colframe=blue!40!white,
    left=1mm, right=1mm, top=1mm, bottom=1mm,
    width=\textwidth,
    enhanced,
    attach boxed title to top center={yshift=-2mm},
    boxed title style={colback=blue!40!white},
    label={box:basesystemprompt}
]
You are a competitive game player. Make sure you read the game instructions carefully, and always follow the required format.
\end{tcolorbox}

\newpage

\subsection{Imperfect Information Games}

\begin{figure}[H]
\centering

\begin{tcolorbox}[
    colback=blue!5!white,
    colframe=blue!40!white,
    left=0.8mm, right=0.8mm, top=0.8mm, bottom=0.8mm,
    boxsep=0.8mm,
    width=\textwidth,
    enhanced,
    breakable,
    attach boxed title to top center={yshift=-2mm},
    boxed title style={colback=blue!40!white},
    title={KuhnPoker Game Starting Prompt},
    fontupper=\footnotesize,
    fonttitle=\footnotesize,
    label=box:KuhnPoker
]
You are Player 0 in a 3 round game of Kuhn Poker.\\
Game Rules:\\
- Kuhn Poker uses a 3-card deck with J, Q, K (J lowest, K highest)\\
- Each player antes 1 chip and receives 1 card each round (note that the cards are dealt without replacement, so you cannot have the same card as your opponent).\\
- Game continues for 3 rounds\\
- The player with the most chips after all rounds wins\\
Action Rules:\\
- '[check]': Pass without betting (only if no bet is on the table)\\
- '[bet]': Add 1 chip to the pot (only if no bet is on the table)\\
- '[call]': Match an opponent's bet by adding 1 chip to the pot\\
- '[fold]': Surrender your hand and let your opponent win the pot\\
\#\#\# Starting round 1 out of 3 rounds. Your card is: 'Q'\\
Player 1, submitted move: '[bet]'.\\
Your available actions are: '[fold]', '[call]'\\
\end{tcolorbox}

\caption{KuhnPoker Game Starting Prompt} \label{box:KuhnPoker}
\end{figure}

\begin{figure}[H]
\centering

\begin{tcolorbox}[
    colback=blue!5!white,
    colframe=blue!40!white,
    left=0.8mm, right=0.8mm, top=0.8mm, bottom=0.8mm,
    boxsep=0.8mm,
    width=\textwidth,
    enhanced,
    attach boxed title to top center={yshift=-2mm},
    boxed title style={colback=blue!40!white},
    title={Briscola Game Starting Prompt},
    fontupper=\footnotesize,
    fonttitle=\footnotesize,
    label=box:Briscola
]

You are playing Briscola - Player 0.\\
Goal: Win tricks and collect the most points (120 total points in the deck).\\
Card Points: A=11, 3=10, K=4, Q=3, J=2, others=0\\
Card Power: A \textgreater{} 3 \textgreater{} K \textgreater{} Q \textgreater{} J \textgreater{} 7 \textgreater{} 6 \textgreater{} 5 \textgreater{} 4 \textgreater{} 2\\
Trump cards beat non-trump cards regardless of power.\\

Action: '[play X]' where X is the position (1-3) of the card in your hand\\

 Briscola game started! Trump suit: \ding{168} (Trump card: Q\ding{168})\\
 Your hand:\\
  \hspace*{1em}1. J\ding{171} [2 pts]\\
  \hspace*{1em}2. K\ding{168} [4 pts] (TRUMP)\\
  \hspace*{1em}3. A\ding{169} [11 pts]\\

No cards played yet this trick.\\

Scores: Player 0: 0 pts | Player 1: 0 pts\\
Trump suit: \ding{168} | Cards left in deck: 34\\

Play a card using [play X]\\
\end{tcolorbox}

\caption{Briscola Game Starting Prompt} \label{box:Briscola}
\end{figure}

\subsection{Negotiation Games}

\begin{figure}[H]
\centering  
\begin{tcolorbox}[
    colback=blue!5!white,
    colframe=blue!40!white,
    left=0.8mm, right=0.8mm, top=0.8mm, bottom=0.8mm,
    boxsep=0.8mm,
    width=\textwidth,
    enhanced,
    breakable,
    attach boxed title to top center={yshift=-2mm},
    boxed title style={colback=blue!40!white},
     title={SimpleNegotiation Game Starting Prompt},
    fontupper=\footnotesize,
    fonttitle=\footnotesize,
    label=box:simpleNegotiation
]
You are Player 0 in the Negotiation Game.\\
You have some resources, and your task is to trade such that the total value of your resources increases.\\
The resources and associated values you currently have are:\\
        \hspace*{2em}+ [Wheat]   Qty: 14   Value: 6\\
        \hspace*{2em}+ [Wood]    Qty: 15   Value: 11\\
        \hspace*{2em}+ [Sheep]   Qty: 18   Value: 18\\
        \hspace*{2em}+ [Brick]   Qty: 12   Value: 27\\
        \hspace*{2em}+ [Ore]     Qty: 22   Value: 38\\
At each turn, you can talk to your opponent and make a trade offer.\\
Use the following special tokens for actions:\\
  \hspace*{1em}- '[Offer: 3 Sheep, 2 Ore -\textgreater{} 5 Brick, 2 Sheep]': [Offer: Offered Resources -\textgreater{} Requested Resources]\\
  \hspace*{1em}- '[Accept]': To accept an incoming offer.\\
  \hspace*{1em}- '[Deny]': To deny an incoming offer (default).\\
The game lasts for 10 turns in total.\\
\end{tcolorbox}

\caption{SimpleNegotiation Game Starting Prompt} \label{fig:simpleNegotiation}
\end{figure}

\begin{figure}[H]
\centering
\begin{tcolorbox}[
    colback=blue!5!white,
    colframe=blue!40!white,
    left=0.8mm, right=0.8mm, top=0.8mm, bottom=0.8mm,
    boxsep=0.8mm,
    width=\textwidth,
    enhanced,
    breakable,
    attach boxed title to top center={yshift=-2mm},
    boxed title style={colback=blue!40!white},
    title={TwoDollar Game Starting Prompt},
    fontupper=\footnotesize,
    fonttitle=\footnotesize,
    label=box:twoDollar
]
[GAME] TWO DOLLAR NEGOTIATION GAME\\
You are Player 0 negotiating with another player over how to split \$2.00 total.
There are 20 maximum rounds.\\
GAME RULES:\\
- You must reach an agreement on how to split the \$2.00\\
- If no agreement is reached, both players get \$0.00\\
- You have secret instructions that only you can see (below)\\
REQUIRED ACTION FORMAT:\\
Always provide your reasoning/persuasion BEFORE the bracketed action.\\
Any text after the bracketed action will be ignored.\\
Available actions:\\
- Make a proposal: ``I think this is fair because... [Propose] \$X.XX"\\
- Accept current proposal: ``This works for me because... [Accept]"\\
- Reject current proposal: ``I can't accept this because... [Reject]"\\
=== YOUR SECRET INSTRUCTIONS ===\\
This person you are dealing with probably will not want to tell you about it but he or she has had a terribly hard month. This past month was full of disaster: money troubles, family difficulties, and none of it your colleague's fault. Through it all, your colleague has been bearing up.\\
Nevertheless, this is zero-sum bargaining, so get all you can. Make up whatever story you wish, and get as much of the \$2.00 as possible. (Just remember that your colleague has been having a very hard time.)\\
VICTORY CONDITION: Reach any deal while being appropriately sympathetic\\
FAILURE CONDITION: No deal reached OR be too harsh given their difficulties\\
{[}GAME{]} === ROUND 1 of 20 ===\\
\end{tcolorbox}

\caption{TwoDollar Game Starting Prompt} \label{box:twoDollar}
\end{figure}

\subsection{Perfect Information Games}

\begin{figure}[H]
\centering

\begin{tcolorbox}[
    colback=blue!5!white,
    colframe=blue!40!white,
    left=0.8mm, right=0.8mm, top=0.8mm, bottom=0.8mm,
    boxsep=0.8mm,
    width=\textwidth,
    enhanced,
    breakable,
    attach boxed title to top center={yshift=-2mm},
    boxed title style={colback=blue!40!white},
    title={SimpleTak Game Starting Prompt},
    fontupper=\footnotesize,
    fonttitle=\footnotesize,
    label=box:SimpleTak
]
You are Player 0 in SimpleTak.\\
On the board, your stones appear as 'O' and your opponent's stones appear as 'X'.\\

On your turn, choose one empty cell (by its numbered index) and place your stone there.
For example, '[12]' places your stone in cell 12.\\

Your objective is to form a continuous path of your stones that connects two opposite edges of the board (top-to-bottom or left-to-right).\\

Current Board:\\

\begin{verbatim}
+----+----+----+----+
| 0  | 1  | 2  | 3  |
+----+----+----+----+
| 4  | 5  | 6  | 7  |
+----+----+----+----+
| 8  | 9  | 10 | 11 |
+----+----+----+----+
| 12 | 13 | 14 | 15 |
+----+----+----+----+
\end{verbatim}
Available Moves: [0], [1], [2], [3], [4], [5], [6], [7], [8], [9], [10], [11], [12], [13], [14], [15]\\
\end{tcolorbox}

\caption{SimpleTak Game Starting Prompt} \label{box:SimpleTak}
\end{figure}

\section{Experimental Setup and Baseline Details}
\label{appendix: Training_settings}

We incorporate three prompt optimization methods to refine prompts using tournament trajectories. Specifically, we leverage offline trajectories collected during the tournament’s self-play process to improve the agents’ prompts. The experimental settings are as follows: the number of generations is set to $5$, the population size to $8$, the number of self-play rounds to $25$, and the number of evaluation rounds to $25$. 
We discuss \texttt{TextGrad} in detail in Section~\ref{app:baseline:Textgrad}, describe our implementation of \texttt{MIPRO} in Section~\ref{app:baseline:mipro}, and provide a comprehensive overview of \texttt{GEPA} in Section~\ref{app:baseline:gepa}. Training details for \texttt{UnstableBaseline} are presented in Section~\ref{app:baseline:unstable_baseline}.

\subsection{Textgrad}
\label{app:baseline:Textgrad}
\textbf{TextGrad}~\citep{yuksekgonul2024textgrad} is a framework that performs "text differentials" to optimize prompts. Within this framework, a text-based loss function analyzes errors, which are then back-propagated to the original prompt through the \texttt{TextGrad} engine. In our case, the goal is to optimize the system prompt of the agent using the trajectories generated under the current system prompt. We design a text-based loss that highlights deficiencies in the generated trajectories. The \texttt{TextGrad} backpropagation engine then propagates gradients back to the system prompt, updating it accordingly. The loss template we adopt is shown in Figure~\ref{fig:Textgrad_loss_template}. 

For each optimization step, we concatenate multiple trajectories, embed them into the template, and use the completed template as the loss input. To ensure balanced feedback, we select an equal number of win, loss, and draw trajectories. This design allows the \texttt{Textgrad} engine to develop a more comprehensive understanding of the current system prompt’s overall game-play patterns.


\begin{center}
\scriptsize   
\resizebox{0.90\textwidth}{!}{
\begin{tabular}{lccccc}
    \toprule
    \multirow{2}{*}{\textbf{Textgrad}} &
    \multicolumn{2}{c}{\textbf{Negotiation}} &
    \multicolumn{2}{c}{\textbf{Imperfect Info}} &
    \multicolumn{1}{c}{\textbf{Perfect Info}} \\
    \cmidrule(lr){2-3}\cmidrule(lr){4-5}\cmidrule(lr){6-6}
     & \textbf{SimpleNegotiation} & \textbf{TwoDollar} & \textbf{KuhnPoker} & \textbf{Briscola} & \textbf{Simpletak} \\
    \midrule
    \multicolumn{6}{l}{\textbf{GPT-4o-mini}} \\
    \midrule
    Trial 1 & 41.3\% & 48.3\% & 58.7\% & 1.3\% & 25.3\% \\
    Trial 2 & 44.7\% & 41.3\% & 56.0\% & 2.0\% & 23.3\% \\
    Trial 3 & 40.0\% & 44.0\% & 52.0\% & 18.0\% & 22.0\% \\
    \midrule
    Avg.    & 42.0\% & 44.6\% & 55.6\% & 7.1\% & 23.6\% \\
    Std.    & 2.4 & 3.5 & 3.4 & 9.4 & 1.7 \\
    \midrule
    \multicolumn{6}{l}{\textbf{Qwen2.5-7B-Instruct}} \\
    \midrule
    Trial 1 & 40.0\% & 38.0\% & 51.3\% & 3.3\% & 18.0\% \\
    Trial 2 & 34.0\% & 34.0\% & 54.7\% & 16.7\% & 22.7\% \\
    Trial 3 & 37.3\% & 16.0\% & 52.7\% & 1.3\% & 26.7\% \\
    \midrule
    Avg.    & 37.1\% & 29.3\% & 52.8\% & 7.1\% & 22.4\% \\
    Std.    & 3.0 & 11.7 & 1.7 & 8.3 & 4.3 \\
    \bottomrule
\end{tabular}
}
\captionof{table}{Performance of the \texttt{Textgrad} method across three independent trials using GPT-4o-mini and Qwen2.5-7B-Instruct. Results are reported as mean win rates with standard deviations.}
\label{tab:Textgrad-results}
\end{center}

\begin{figure}[ht]
\centering

\begin{tcolorbox}[
    colback=blue!5!white,
    colframe=blue!40!white,
    left=1mm, right=1mm, top=1mm, bottom=1mm,
    width=\textwidth,
    enhanced,
    attach boxed title to top center={yshift=-2mm},
    boxed title style={colback=blue!40!white},
    title={Text-based loss template for Textgrad},
    label=box:textgrad_loss
]
You are an objective evaluator for a two-player zero-sum game agent's SYSTEM PROMPT. \\

Goal of the SYSTEM PROMPT (what it MUST enforce): \\
- Maximize the agent's win rate. \\
- Minimize the opponent's win rate. \\
- Have strategies that lead to victory. \\
- Ensure all moves strictly follow game rules and formats. \\

Here are some game trajectories using the current SYSTEM PROMPT: \\
\{\{trajectory examples\}\} \\

Identify specific weaknesses or flaws in the SYSTEM PROMPT that may have led to losses or suboptimal plays. \\
Do NOT suggest improvements or rewrites, only identify weaknesses. \\
Be very concise and specific.
\end{tcolorbox}

\caption{Text-based loss template for Textgrad} \label{fig:Textgrad_loss_template}
\end{figure}

\subsection{MIPRO} \label{app:baseline:mipro}
\textbf{MIPRO} \citep{opsahl2024optimizing} optimizes prompts based on downstream task performance. In our work, we adopt the \texttt{MIPROv2} implementation provided by the \texttt{Dspy} library ~\citep{khattab2023dspy}. The optimization procedure consists of three main steps: (1) Sampling examples: For each candidate prompt, \texttt{MIPRO} samples a set of examples. (2) Proposing prompts: New system prompts are proposed by a propose model based on the current system prompt, along with additional game-related information such as the program description, data description, random sampling tips, and few-shot examples. (3) Evaluation through trials: Several trials are conducted to evaluate which combination of proposed prompts and few-shot examples yields the best performance. A Bayesian search strategy is then applied to guide the selection of the next candidate combination, improving efficiency and reducing computational cost. 

In our experiments, we only have access to offline game data. Therefore, we treat each step in a trajectory as an individual data point. For each step, we record the outcome (win, loss, or draw) of the trajectory it belongs to. \texttt{MIPRO}’s evaluation metric is defined based on the model’s re-inference of these steps: (1) If the model outputs an invalid action (i.e., one that does not conform to the required format), the score is $0$. (2) For steps from winning trajectories, if the model predicts the same action as the original step, the score is $1$; otherwise, it is $0$. (3) For steps from losing trajectories, if the model predicts the same action, the score is $0$ (to discourage repeating losing moves); otherwise, it is $1$. (4) For steps from draw trajectories, if the model predicts the same action, the score is $0.2$; otherwise, it is $0.5$, encouraging exploration beyond draw-inducing moves.

This scoring scheme encourages the model to replicate winning strategies, avoid losing ones, and explore alternatives to drawn outcomes. The overall \texttt{MIPRO} scoring standard is shown in Figure~\ref{fig:mipro_scoring_standard}. In practice, we set the number of proposed prompts to $6$, the number of few-shot examples to $3$, and the number of trials to $10$. If the optimal configuration includes few-shot examples, these are appended to the final proposed system prompt to form the new system prompt.

\begin{figure}[ht]
\centering

\begin{tcolorbox}[
    colback=blue!5!white,
    colframe=blue!40!white,
    left=1mm, right=1mm, top=1mm, bottom=1mm,
    width=\textwidth,
    enhanced,
    attach boxed title to top center={yshift=-2mm},
    boxed title style={colback=blue!40!white},
    title={MIPRO scoring standard},
    label=box:mipro_scoring
]

\textbf{Invalid Action}:
score = 0.0 \\

\textbf{Win Trajectory}:
Action match: score = 1.0 / 
Action mismatch: score = 0.0 \\

\textbf{Lose Trajectory}: 
Action match: score = 0.0 /
Action mismatch: score = 1.0 \\

\textbf{Draw Trajectory}: 
Action match: score = 0.2 /
Action mismatch: score = 0.5

\end{tcolorbox}

\caption{MIPRO scoring standard} \label{fig:mipro_scoring_standard}
\end{figure}

\begin{center}
\scriptsize   
\resizebox{0.90\textwidth}{!}{
\begin{tabular}{lccccc}
\toprule
\multirow{2}{*}{\textbf{MIPRO}} &
\multicolumn{2}{c}{\textbf{Negotiation}} &
\multicolumn{2}{c}{\textbf{Imperfect Info}} &
\multicolumn{1}{c}{\textbf{Perfect Info}} \\
\cmidrule(lr){2-3}\cmidrule(lr){4-5}\cmidrule(lr){6-6}
 & \textbf{SimpleNegotiation} & \textbf{TwoDollar} & \textbf{KuhnPoker} & \textbf{Briscola} & \textbf{Simpletak} \\
\midrule
\multicolumn{6}{l}{\textbf{GPT-4o-mini}} \\
\midrule
Trial 1 & 38.7\% & 53.3\% & 50.7\% & 23.3\% & 16.0\% \\
Trial 2 & 38.0\% & 52.7\% & 60.0\% & 32.7\% & 20.0\% \\
Trial 3 & 38.7\% & 46.7\% & 54.7\% & 3.33\% & 21.3\% \\
\midrule
Avg. & 38.4\%  & 50.9\% & 55.1\% & 19.7\% & 19.1\% \\
Std. & 0.38 & 3.67 & 4.68 & 14.99 & 2.78 \\
\midrule
\multicolumn{6}{l}{\textbf{Qwen2.5-7B-Instruct}} \\
\midrule
Trial 1 & 43.3\% & 40.7\% & 54.0\% & 2.0\% & 18.7\% \\
Trial 2 & 37.3\% & 52.0\% & 50.0\% & 2.0\% & 19.3\% \\
Trial 3 & 46.7\% & 50.0\% & 57.3\% & 2.7\% & 24.7\% \\
\midrule
Avg. & 42.4\%  & 47.5\% & 53.8\%  & 2.2\% & 20.9\% \\
Std. & 4.73 & 6.05 & 3.67 & 0.38 & 3.29 \\
\midrule
\bottomrule
\end{tabular}
}
\captionof{table}{Performance of the \texttt{MIPRO} method across three independent trials using GPT-4o-mini and Qwen2.5-7B-Instruct. Results are reported as mean win rates with corresponding standard deviations.}
\label{tab:MIPRO-results}
\end{center}

\subsection{GEPA} \label{app:baseline:gepa}
\textbf{GEPA} ~\citep{agrawal2025gepa} builds upon the high-level idea of \texttt{MIPRO}, but extends it by incorporating both evaluation scores and explicit feedback from the evaluation metric to guide prompt optimization. The process can be summarized as follows: (1) Initial evaluation: Run a set of examples through the evaluation metric to obtain an initial score and feedback. (2) Prompt proposal: Generate a new prompt based on the current prompt and the feedback collected. (3) Testing and retention: Evaluate the new prompt on a mini-batch. If its score surpasses the initial score, retain it in the candidate pool. (4) Candidate selection: In the next round, apply a Pareto-based filtering strategy to identify the set of candidate prompts that dominate on the validation set. Select one of these Pareto-optimal prompts for further iteration. (5) Stopping condition: The optimization continues until the maximum number of evaluation metric calls reaches a predefined limit. 

In our experiments, we set the maximum number of evaluation metric calls to $100$ for each prompt optimization in \texttt{GEPA}. For win and lose trajectories, we adopt the same evaluation metric as \texttt{MIPRO}. For draw trajectories, we assign a score of $0$ when the predicted action matches the trajectory action, and a score of $1$ otherwise. In addition, we incorporate feedback signals in \texttt{GEPA} evaluation metric. The structured feedback template shown in Figure~\ref{fig:gepa_scoring_standard} is used during \texttt{GEPA} evaluation.

    \begin{center}
    \scriptsize   
    \resizebox{0.90\textwidth}{!}{
    \begin{tabular}{lccccc}
    \toprule
    \multirow{2}{*}{\textbf{GEPA}} &
    \multicolumn{2}{c}{\textbf{Negotiation}} &
    \multicolumn{2}{c}{\textbf{Imperfect Info}} &
    \multicolumn{1}{c}{\textbf{Perfect Info}} \\
    \cmidrule(lr){2-3}\cmidrule(lr){4-5}\cmidrule(lr){6-6}
     & \textbf{SimpleNegotiation} & \textbf{TwoDollar} & \textbf{KuhnPoker} & \textbf{Briscola} & \textbf{Simpletak} \\
    \midrule
    \multicolumn{6}{l}{\textbf{GPT-4o-mini}} \\
    \midrule
    Trial 1 & 34.7\% & 32.7\% & 54.7\% & 1.3\% & 23.3\% \\
    Trial 2 & 38.0\% & 43.3\% & 50.7\% & 3.3\% & 29.3\% \\
    Trial 3 & 38.0\% & 45.3\% & 51.3\% & 5.3\% & 28.0\% \\
    \midrule
    Avg. & 36.8\%             & 40.4\% & 52.2\%            & 3.3\% & 26.9\% \\
    Std. & 1.92 & 6.81 & 2.14 & 2.00 & 3.15 \\
    \midrule
    \multicolumn{6}{l}{\textbf{Qwen2.5-7B-Instruct}} \\
    \midrule
    Trial 1 & 29.3\% & 22.7\% & 56.0\% & 4.0\% & 20.0\% \\
    Trial 2 & 38.7\% & 30.0\% & 54.0\% & 2.0\% & 12.0\% \\
    Trial 3 & 35.3\% & 42.7\% & 57.3\% & 2.0\% & 26.0\% \\
    \midrule
    Avg. & 34.4\%  & 31.7\% & 55.8\%  & 3.3\% & 19.3\% \\
    Std. & 4.73 & 10.12 & 1.68 & 1.55 & 7.02 \\
    \bottomrule
    \end{tabular}
    }
    \captionof{table}{Performance of the \texttt{GEPA} method across three independent trials using GPT-4o-mini and Qwen2.5-7B-Instruct. Results are reported as mean win rates with corresponding standard deviations.}
    \label{tab:GEPA-results}
\end{center}

\begin{figure}[H]
\centering

\begin{tcolorbox}[
    colback=blue!5!white,
    colframe=blue!40!white,
    left=1mm, right=1mm, top=1mm, bottom=1mm,
    width=\textwidth,
    enhanced,
    attach boxed title to top center={yshift=-2mm},
    boxed title style={colback=blue!40!white},
    title={GEPA scoring standard},
    label=box:gepa_scoring
]
\# invalid action \\
score = 0.0 \\
feedback = ``Your predicted action is invalid. Please ensure that your action is a valid move in the game. Here is the reasoning process \{\{model\_raw\_output\}\}. Think about how you could have reasoned to choose a valid action that leads to a WIN."\\

\# Win Trajectory \\
\# Action match \\
score = 1.0 \\
feedback = ``You correctly predicted the action \{\{pred\_action\}\} that led to a WIN. This action was indeed the one taken in the winning trajectory. Great job!" \\

\# Action mismatch \\
score = 0.0 \\
feedback = ``You predicted the action \{\{pred\_action\}\}, but the action taken in the winning trajectory was \{\{traj\_action\}\}. This mismatch means you did not predict the winning action correctly. Here is the reasoning process \{\{pred\_raw\_action\}\}. Think about how you could have reasoned to get the correct action." \\

\# Lose Trajectory \\
\# Action match \\
score = 0.0 \\
feedback = ``You correctly predicted the action \{\{pred\_action\}\} that led to a LOSE. However, this action was part of a losing trajectory. While your prediction matches the trajectory, it did not lead to a win. Here is the reasoning process \{\{pred\_raw\_action\}\}. Think about how you could have reasoned to choose an action that leads to a WIN." \\

\# Action mismatch \\
score = 1.0 \\
feedback = ``You predicted the action \{\{pred\_action\}\}, but the action taken in the losing trajectory was \{\{traj\_action\}\}. This mismatch means you did not predict the losing action correctly. Here is the reasoning process \{\{pred\_raw\_action\}\}. Think about how you could have reasoned to choose an action that leads to a WIN." \\

\# Draw Trajectory \\
\# Action match \\
score = 0.0 \\
feedback = ``You predicted the action \{\{pred\_action\}\}, which matches the action taken in the TIE trajectory. However, since the trajectory resulted in a TIE, this does not help in achieving a WIN. Here is the reasoning process \{\{pred\_raw\_action\}\}. Think about how you could have reasoned to choose an action that leads to a WIN." \\

\# Action mismatch \\
score = 1.0 \\
feedback = ``You predicted the action \{\{pred\_action\}\}, but the action taken in the TIE trajectory was \{\{traj\_action\}\}. This mismatch means you did not predict the TIE action correctly. Here is the reasoning process \{\{pred\_raw\_action\}\}. Think about how you could have reasoned to choose an action that leads to a WIN." \\

\end{tcolorbox}

\caption{GEPA scoring standard} \label{fig:gepa_scoring_standard}
\end{figure}

\subsection{UnstableBaseline} \label{app:baseline:unstable_baseline}
\textbf{UnstableBaseline}~\citep{Guertler_UnstableBaselines_2025} is an asynchronous online multi-agent reinforcement learning library that uses Low-Rank Adapters (LoRA) for model training. Unlike its peers such as \texttt{Verifiers}~\citep{brown_verifiers_2025} and \texttt{SPIRAL}~\citep{liu2025spiral}, \texttt{UnstableBaseline} is designed to be lightweight and closely integrated with the \texttt{TextArena}~\citep{guertler2025textarena} environment, in the same spirit that the baseline~\citep{baselines} library complements OpenAI Gym~\citep{1606.01540}. 

For our experiments, we used the default training configuration provided by \texttt{UnstableBaseline} without additional hyperparameter tuning.
Specifically, we trained Qwen2.5-7B-Instruct with LoRA adapters applied to the attention and feedforward projections, using rank $r=16$, $\alpha=32$, and dropout $=0.0$.
Training was performed using the REINFORCE algorithm~\citep{Williams:92}.

From the best performing checkpoints, we held $3$ rounds of $50$ games against each of our evaluation models that is similarly used in our training settings for the other prompt evolution experiments. Their results can be found in table \ref{tab:unstable-results}.

\begin{center}
\vspace{0.7pt}
\resizebox{0.90\textwidth}{!}{
\begin{tabular}{lccccc}
    \toprule
    \multirow{2}{*}{\textbf{UnstableBaseline}} &
    \multicolumn{2}{c}{\textbf{Negotiation}} &
    \multicolumn{2}{c}{\textbf{Imperfect Info}} &
    \multicolumn{1}{c}{\textbf{Perfect Info}} \\
    \cmidrule(lr){2-3}\cmidrule(lr){4-5}\cmidrule(lr){6-6}
     & \textbf{SimpleNegotiation} & \textbf{TwoDollar} & \textbf{KuhnPoker} & \textbf{Briscola} & \textbf{Simpletak} \\
    \midrule
    \multicolumn{6}{l}{\textbf{Qwen2.5-7B-Instruct}} \\
    \midrule
    Gemini-2.5-Flash-Lite & 54.7\% & 43.3\% & 50.0\% & 88.6\% & 90.0\% \\
    Grok-4-Fast-Non-Reasoning  & 44.7\% & 22.0\% & 54.7\% & 33.3\% & 20.0\% \\
    Qwen3-235B-A22B-Instruct-2507 & 24.0\% & 26.0\% & 53.3\% & 38.0\% & 32.0\% \\
    \midrule
    Avg.    & 41.1\% & 30.4\% & 52.6\% & 53.3\% & 47.3\% \\
    Std.    & 15.6 & 11.3 & 2.40 & 30.7 & 37.4 \\
    \bottomrule
\end{tabular}
}
\captionof{table}{Performance of the \texttt{UnstableBaseline} method across three independent trials using Qwen2.5-7B-Instruct. Results are reported as mean win rates with corresponding standard deviations, where each mean win rate was from the average of 3 rounds of 50 matches with each opponent, with alternating starting positions.}
\label{tab:unstable-results}
\end{center}

\subsection{SPIRAL}
\label{app:baseline:spiral}
\textbf{SPIRAL}~\citep{liu2025spiral} is a framework that enables language models to autonomously develop reasoning capabilities through self-play in multi-turn, zero-sum games. For our experiments, we train Qwen2.5-7B-Instruct using Reinforce, following the default rollout size in the provided example, each rollout comprising 128 games over 400 total steps. We then select the best-performing checkpoint and evaluate it over three rounds of 50 games each.

\section{Comparison with Existing Prompt Optimization Methods} \label{app:section:prompt_optimization_comparison}

In Section~\ref{appendix: Training_settings}, we introduced three baseline prompt optimization methods. Here, we further highlight how our approach differs from these methods.  

As shown in Figure~\ref{fig:methodology}, our method evolves a population of prompts using elitism, local edits/expansions, random exploration, and memory-augmented updates. Random exploration enables broader search over prompt variants, while memory-augmented updates leverage insights distilled from self-play trajectories to refine new prompt candidates.  

\textbf{Versus TextGrad.} \texttt{TextGrad} relies on hand-crafted text losses and gradient-style backpropagation over natural language. In contrast, our method is entirely \emph{gradient-free}: it requires no differentiable loss functions or template engineering. This avoids sensitivity to wording in loss templates and reduces dependence on diagnostic outputs, where weak language models often fail to generate meaningful diagnostic responses.  

\textbf{Versus MIPRO.} \texttt{MIPRO} frames optimization as Bayesian search over (prompt, few-shot) pairs, requiring many trials and frequent evaluation metric calls. Its effectiveness hinges on having a well-defined evaluation metric, which is difficult to obtain in text-based games where no concise supervision signal exists. As a result, \texttt{MIPRO} consumes many tokens without achieving strong performance. Our method, by contrast, does not rely on explicit evaluation metrics. It can leverage diverse signals from self-play trajectories, achieving better performance with fewer model calls and without heavy trial scheduling. 

\textbf{Versus GEPA.} \texttt{GEPA} extends \texttt{MIPRO}'s evaluation process by augmenting it with verbose textual feedback and repeatedly querying an evaluation oracle until its call budget is exhausted, making it heavily dependent on the quality of the evaluation metric. Its key mechanism is a Pareto-based selection strategy, which identifies promising prompts from the candidate pool based on the Pareto frontier. However, the construction of this frontier relies strongly on the evaluation scores, and when the metric is not well-defined, the selected prompts may not be optimal. In contrast, our method replaces such reliance on external feedback with \emph{memory-augmented edits} distilled directly from self-play outcomes, while maintaining diversity through randomization. This design reduces token usage, improves robustness under noisy feedback, and removes dependence on external evaluation metrics.  

\section{Token Cost Comparison}
\label{appendix:token_cost}

\begin{table}[!ht]
    \centering
    \scriptsize
    \setlength{\tabcolsep}{0.1cm}
    \resizebox{0.7\textwidth}{!}{
    \begin{tabular}{lcccc}
    \toprule
    \textbf{Optimizer} & \textbf{SimpleNegotiation} & \textbf{KuhnPoker} & \textbf{SimpleTak} & \textbf{Avg. Tokens} \\
    \midrule
    Textgrad  & 842        & 986        & 938        & 922 \\
    MIPRO     & 145{,}864  & 162{,}084  & 754{,}534  & 354{,}161 \\
    GEPA      & 110{,}325  & 119{,}365  & 111{,}907  & 113{,}865 \\
    \textbf{\NAME{} (Ours)} & 87{,}364   & 94{,}160   & 89{,}152   & 90{,}575 \\
    \bottomrule
    \end{tabular}
    }
    \caption{Output token cost for each prompt optimization method (exact counts).}
    \label{tab:token-cost}
\end{table}

\section{Full Results}
\label{appendix: Full_Results}

\begin{center}
    \scriptsize   
    \resizebox{0.90\textwidth}{!}{
    \begin{tabular}{lccccc}
    \toprule
    \multirow{2}{*}{\textbf{\NAME{}}} &
    \multicolumn{2}{c}{\textbf{Negotiation}} &
    \multicolumn{2}{c}{\textbf{Imperfect Info}} &
    \multicolumn{1}{c}{\textbf{Perfect Info}} \\
    \cmidrule(lr){2-3}\cmidrule(lr){4-5}\cmidrule(lr){6-6}
     & \textbf{SimpleNegotiation} & \textbf{TwoDollar} & \textbf{KuhnPoker} & \textbf{Briscola} & \textbf{Simpletak} \\
    \midrule
    \multicolumn{6}{l}{\textbf{GPT-4o-mini}} \\
    \midrule
    Trial 1 & 57.3\% & 46.0\% & 54.0\% & 54.0\% & 45.3\% \\
    Trial 2 & 55.3\% & 62.7\% & 57.3\% & 38.0\% & 40.7\% \\
    Trial 3 & 52.0\% & 48.7\% & 55.3\% & 36.0\% & 39.3\% \\
    \midrule
    Avg. & 54.9\% & 52.4\% & 55.6\% & 42.7\% & 41.8\% \\
    Std. & 2.69 & 8.95 & 1.68 & 9.87 & 3.15\\
    \midrule
    \multicolumn{6}{l}{\textbf{Qwen2.5-7B-Instruct}} \\
    \midrule
    Trial 1 & 48.0\% & 53.3\% & 60.7\% & 41.3\% & 37.3\% \\
    Trial 2 & 47.3\% & 54.0\% & 59.3\% & 26.0\% & 32.0\% \\
    Trial 3 & 48.7\% & 38.0\% & 60.0\% & 27.3\% & 41.3\% \\
    \midrule
    Avg. & 48.0\%  & 48.4\% & 60.0\%  & 31.5\% & 36.9\% \\
    Std. & 0.67 & 9.05 & 0.67 & 8.49 & 4.68 \\
    \bottomrule
    \end{tabular}
    }
    \captionof{table}{Performance of the \NAME{} method across three independent trials using GPT-4o-mini and Qwen2.5-7B-Instruct. Results are reported as mean win rates with corresponding standard deviations.}
    \label{tab:main-results}
\end{center}
\vspace{0.7pt}

\begin{center}
    \scriptsize   
    \resizebox{0.90\textwidth}{!}{
    \begin{tabular}{lccccc}
    \toprule
    \multirow{2}{*}{\textbf{\NAME{}}} &
    \multicolumn{2}{c}{\textbf{Negotiation}} &
    \multicolumn{2}{c}{\textbf{Imperfect Info}} &
    \multicolumn{1}{c}{\textbf{Perfect Info}} \\
    \cmidrule(lr){2-3}\cmidrule(lr){4-5}\cmidrule(lr){6-6}
     & \textbf{SimpleNegotiation} & \textbf{TwoDollar} & \textbf{KuhnPoker} & \textbf{Briscola} & \textbf{Simpletak} \\
    \midrule
    \multicolumn{6}{l}{\textbf{GPT-4o-mini}} \\
    \midrule
    Gemini-
2.5-Flash-Lite  & 94.0\% & 46.7\% & 56.0\% & 55.3\% & 62.0\% \\
    Grok-
4-Fast-Non-Reasoning    & 30.7\% & 46.7\% & 58.0\% & 39.3\% & 45.3\% \\
    Qwen3-235B-
A22B-Instruct-2507    & 40.0\% & 64.0\% & 52.7\% & 33.3\% & 18.0\% \\
    \midrule
    Avg. & 54.9\% & 52.4\% & 55.6\% & 42.7\% & 41.8\% \\
    Std. & 34.2 & 10.0 & 2.7 & 11.4 & 22.2 \\
    \midrule
    \multicolumn{6}{l}{\textbf{Qwen2.5-7B-Instruct}} \\
    \midrule
    Gemini-
2.5-Flash-Lite  & 90.7\% & 47.3\% & 58.0\% & 61.3\% & 69.3\% \\
    Grok-
4-Fast-Non-Reasoning    & 21.3\% & 38.0\% & 55.3\% & 18.0\% & 32.7\% \\
    Qwen3-235B-
A22B-Instruct-2507    & 32.0\% & 60.0\% & 66.7\% & 15.3\% & 8.7\% \\
    \midrule
    Avg. & 48.0\%  & 48.4\% & 60.0\%  & 31.5\% & 36.9\% \\
    Std. & 37.3 & 11.0 & 5.9 & 25.8 & 30.6 \\
    \bottomrule
    \end{tabular}
    }
    \captionof{table}{Performance of the \NAME{} method across each opponent model using GPT-4o-mini and Qwen2.5-7B-Instruct. Results are reported as mean win rates with corresponding standard deviations of the win rates across opponent models.}
    \label{tab:main-results-by-opponent}
\end{center}

\vspace{0.7pt}
\begin{table*}[ht]
    \centering
    \resizebox{0.90\textwidth}{!}{
    \begin{tabular}{lccccc|c|c}
    \toprule
    \multirow{2}{*}{\textbf{Optimizer}} &
    \multicolumn{2}{c}{\textbf{Negotiation}} &
    \multicolumn{2}{c}{\textbf{Imperfect Info}} &
    \multicolumn{1}{c}{\textbf{Perfect Info}} &
    \multirow{2}{*}{\begin{tabular}{@{}c@{}}\textbf{Mean}\\\textbf{Win Rate}\end{tabular}} &
    \multirow{2}{*}{\begin{tabular}{@{}c@{}}\textbf{Mean}\\\textbf{RSE}\end{tabular}} \\
    \cmidrule(lr){2-3}\cmidrule(lr){4-5}\cmidrule(lr){6-6}
     & \textbf{SimpleNegotiation} & \textbf{TwoDollar} & \textbf{KuhnPoker} & \textbf{Briscola} & \textbf{SimpleTak} & & \\
    \midrule

    \multicolumn{8}{l}{\textbf{GPT-4o-mini}} \\
    baseline & 31.3\% & 32.2\% & 39.1\% & 0.3\%  & 21.4\% & 25.1\% & 44.9\% \\
    \textbf{\NAME{} (Ours)} & \textbf{54.9\%} & \textbf{52.4\%} & \textbf{55.6\%} & \textbf{42.7\%} & \textbf{41.8\%} & \textbf{49.5\%} & \textbf{6.4\%} \\
    \midrule

    \multicolumn{8}{l}{\textbf{Qwen2.5-7B-Instruct}} \\
    baseline & 24.0\% & 17.1\% & 45.3\% & 2.8\%  & 15.1\% & 20.9\% & 30.1\% \\
    \textbf{\NAME{} (Ours)} & \textbf{48.0\%} & \textbf{48.4\%} & \textbf{60.0\%} & 31.1\% & 34.0\% & 44.3\% & \textbf{6.1\%} \\
    \midrule

    \multicolumn{8}{l}{\textbf{Gemini-2.5-Flash}} \\
    baseline & 14.0\% & 15.0\% & 50.0\% & 32.0\% & 26.0\% & 27.4\% & -\% \\
    \textbf{\NAME{} (Ours)} & \textbf{30.0\%} & \textbf{35.0\%} & \textbf{58.0\%} & \textbf{49.0\%} & \textbf{32.0\%} & \textbf{40.8\%} & -\% \\
    \bottomrule
    \end{tabular}
    }
    \caption{Benchmark results for baseline and \NAME{} across multiple tasks. Each win rate is the mean across three evaluation models (Sec.~\ref{evaluation_setup}). For per-opponent breakdown, refer to Appendix~\ref{tab:main-results-by-opponent}.}
    \label{tab:main-results-with-gemini}
\end{table*}

\section{Game Environments}
\label{app:game_env}
We provide more detailed descriptions of the text-based games selected from \texttt{TextArena}~\citep{guertler2025textarena} and \texttt{SPIN-Bench}~\citep{yao2025spin}.

\medskip
\textbf{Simple Negotiation}~\citep{Nash1950THEBP} requires players to reason about trade-offs through the exchange of resources such as wood, wheat, sheep, brick, and ore. Each player aims to maximize the value of their inventory by making offers and counteroffers with their opponent. Success depends on each player's ability to infer the opponent’s valuation of resources and strategically increase their own portfolio without making disadvantageous trades.

\textbf{Two Dollar Game}~\citep{MIT_OCW_Negotiation2001} is a classroom negotiation game where two players have to agree on how to divide a fixed sum of \$2.00. Typically, players each receive private role instructions that impose certain constraints or encourage specific negotiation styles. This asymmetric information requires players to balance their objectives with compromises while inferring the opponent's position.

\textbf{Kuhn Poker}~\citep{Kuhn1951} is a simplified form of poker played with three cards (Jack, Queen, and King). Two players each receive one card, while the third remains unseen. A single round of betting follows, where players can check, bet, call, or fold. If neither folds, the winner is determined by the higher card.

\textbf{Briscola}~\citep{pagat_briscola} is a traditional Italian trick-taking card game played with a 40-card deck. At the start, a single card is revealed to determine the trump suit, and each player is dealt a hand of cards. Players take turns playing one card per trick, with the highest card of the leading suit or the highest trump winning the round. The objective is to accumulate points by capturing valuable cards, requiring players to balance tactical play with long-term strategy and inference of the opponent’s hand.

\textbf{Simple Tak}~\citep{Rothfuss2011} is a two-player connection game inspired by the traditional game Tak. Players place tiles on a square grid with the objective of forming a continuous path that connects opposite sides of the board. Unlike full Tak, stacking pieces is not allowed, though players may block their opponent’s path by occupying critical spaces. The game emphasizes spatial reasoning, foresight, and the balance between advancing one’s own path and disrupting the opponent’s progress.

\section{Insight Case Analysis}
\label{appendix:insightcaseanalysis}

We analyze the high-quality insights stored in the memory bank across different games. These insights emerge from self-play trajectories and contribute to prompt optimization by encoding transferable strategic knowledge. We identify two primary categories of insights that improve game performance: (1) game-specific strategic principles that capture tactical knowledge, and (2) opponent modeling insights that focus on understanding and responding to other players.

\paragraph{Game-Specific Strategic Principles.}
These insights capture tactical knowledge that helps agents make better in-game decisions. They encode domain-specific heuristics that would otherwise require many episodes to rediscover.

\begin{figure}[H]
\centering
\begin{tcolorbox}[
    colback=blue!5!white,
    colframe=blue!40!white,
    left=1mm, right=1mm, top=1mm, bottom=1mm,
    width=\textwidth,
    enhanced,
    attach boxed title to top center={yshift=-2mm},
    boxed title style={colback=blue!40!white},
    title={Kuhn Poker Strategic Insights},
    label=box:kuhn_poker_strategic
]

\textbf{Insight 1 (Pressure-based betting):}
``In future games, consider a strategy where you bet or call more frequently in early rounds, even with weaker cards, to increase potential pots and apply pressure on the opponent, especially when no initial bets are made."

\textbf{Insight 2 (Hand strength exploitation):}
``In future games, players should adopt a more aggressive betting strategy when holding stronger cards, such as K, to force opponents into tough situations that might lead to folds or allow the player to take control of the pot more effectively."

\end{tcolorbox}
\caption{Kuhn Poker strategic insights. These insights encode betting principles that balance aggression with hand strength, helping agents avoid predictable play patterns while maximizing expected value.}
\label{box:kuhn_poker_strategic}
\end{figure}

\begin{figure}[H]
\centering
\begin{tcolorbox}[
    colback=blue!5!white,
    colframe=blue!40!white,
    left=1mm, right=1mm, top=1mm, bottom=1mm,
    width=\textwidth,
    enhanced,
    attach boxed title to top center={yshift=-2mm},
    boxed title style={colback=blue!40!white},
    title={Briscola Strategic Insights},
    label=box:briscola_strategic
]

\textbf{Insight 1 (Trump timing):}
``In future turns, prioritize using the Ace or trump cards at key moments to control the trick."

\textbf{Insight 2 (Point maximization):}
``When holding a trump card, prioritize using it to capture high-point non-trump cards led by the opponent, especially during the mid-game when more point cards are likely to be in play. This maximizes point gain and helps secure early leads."

\end{tcolorbox}
\caption{Briscola strategic insights. These insights capture the timing and resource allocation principles for trump cards, enabling agents to maximize point capture rather than using high-value cards indiscriminately.}
\label{box:briscola_strategic}
\end{figure}

Strategic principles reduce the search space for decision-making by providing domain-appropriate heuristics. Rather than exploring all possible actions uniformly, agents can prioritize moves that align with proven tactical patterns, leading to faster convergence and more consistent performance.

\paragraph{Opponent Modeling and Negotiation Dynamics.}
These insights focus on understanding opponent behavior and leveraging psychological or structural aspects of multi-agent interactions.

\begin{figure}[H]
\centering
\begin{tcolorbox}[
    colback=blue!5!white,
    colframe=blue!40!white,
    left=1mm, right=1mm, top=1mm, bottom=1mm,
    width=\textwidth,
    enhanced,
    attach boxed title to top center={yshift=-2mm},
    boxed title style={colback=blue!40!white},
    title={Simple Negotiation Insights},
    label=box:simple_neg_insights
]

\textbf{Insight 1 (Preference inference):}
``To improve negotiation outcomes, Player should analyze the resource preferences of other Player more closely and tailor offers to match those preferences, possibly by proposing trades that highlight the mutual benefits rather than assuming equal value among resources."

\textbf{Insight 2 (Information gathering):}
``In future negotiations, it would be beneficial to engage in dialogue to better understand Player 1's resource priorities before making trade offers, potentially increasing the chance of acceptance and maximizing resource value."

\end{tcolorbox}
\caption{Simple Negotiation insights. These insights reveal that players have asymmetric resource valuations, a concept not explicitly stated in the game description, and encourage proactive information gathering before committing to offers.}
\label{box:simple_neg_insights}
\end{figure}

\begin{figure}[H]
\centering
\begin{tcolorbox}[
    colback=blue!5!white,
    colframe=blue!40!white,
    left=1mm, right=1mm, top=1mm, bottom=1mm,
    width=\textwidth,
    enhanced,
    attach boxed title to top center={yshift=-2mm},
    boxed title style={colback=blue!40!white},
    title={Two Dollars Negotiation Insight},
    label=box:two_dollars_insight
]

\textbf{Insight (Time pressure leverage):}
``To improve future negotiations, I could clearly convey the importance of reaching an agreement within the limited rounds available, perhaps framing my offer as a time-sensitive opportunity that other player wouldn't want to miss, thereby encouraging a quicker consensus."

\end{tcolorbox}
\caption{Two Dollars insight. This insight captures a negotiation tactic that exploits the finite round structure, encouraging agents to use time pressure as a persuasion mechanism.}
\label{box:two_dollars_insight}
\end{figure}

Opponent modeling insights enable agents to move beyond self-centered optimization toward strategic reasoning that accounts for the other player's objectives and constraints. By understanding that opponents have different preferences or that structural features like round limits can be leveraged, agents can craft more effective proposals and responses. These insights are particularly valuable in negotiation games where success depends on predicting and influencing opponent behavior.
\section{Prompt Case Analysis}
\label{appendix:prompt_analysis}

\begin{figure}[H]
    \centering
\includegraphics[width=0.9\columnwidth]{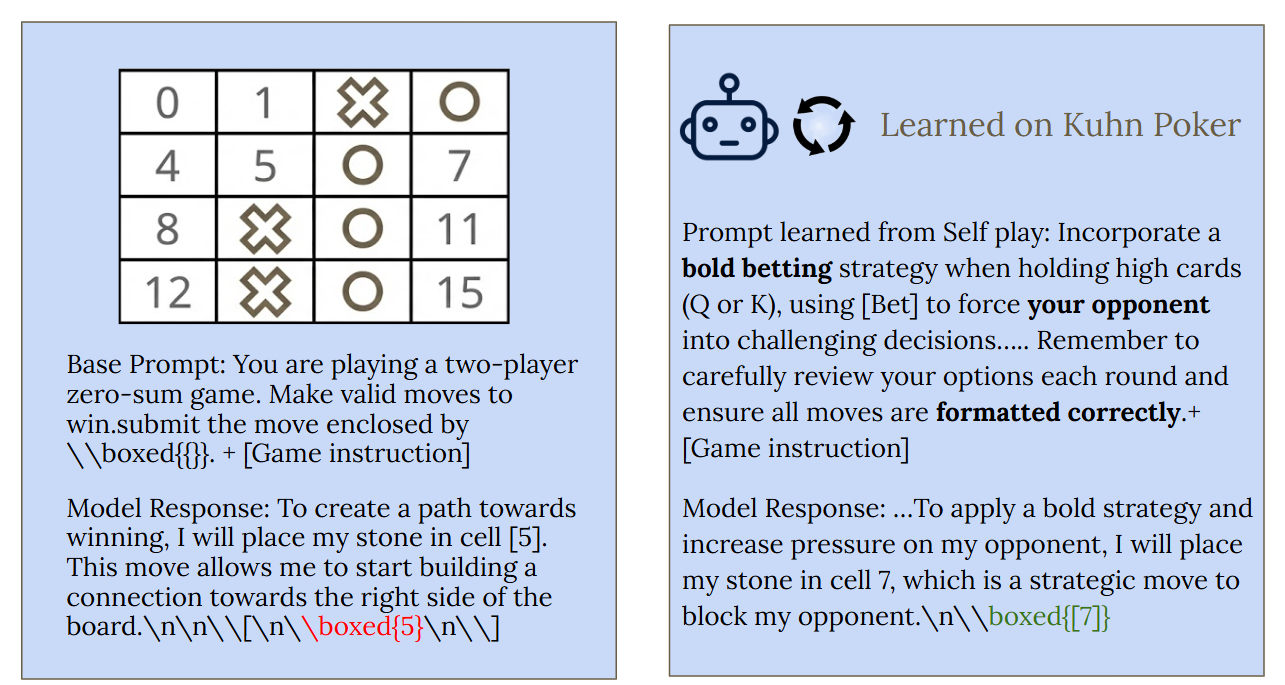}
    \caption{In this example (Fig.~\ref{fig:case_example}), the agent plays \(\times\), while the opponent (\(\circ\)) is one move away from victory. The prompt on the right is learned via self-play in \textsc{Kuhn Poker} and distills transferable behaviors, including opponent modeling, general strategic principles, and strict output-format constraints. Conditioned on this prompt, the agent handles the scenario more reliably; in this short-horizon case, the adapted prompt enables the agent to identify the correct blocking move and prevent the opponent's immediate win.
}
    \label{fig:case_example}
\end{figure}
\begin{figure}[H]
    \centering
\begin{tcolorbox}[
    colback=blue!5!white,
    colframe=blue!40!white,
    left=1mm, right=1mm, top=1mm, bottom=1mm,
    width=\textwidth,
    enhanced,
    attach boxed title to top center={yshift=-2mm},
    boxed title style={colback=blue!40!white},
    title={Simple Negotiation System Prompt Transfer to Simple Tak Example},
    label=box: Simple Negotiation
]

Updated System Prompt:

`` Evaluate Opponent current inventory and preferences before crafting an initial trade proposal. Offer a carefully curated mix of both lower-value and higher-value items that not only maximize your immediate gain but also align with Player 1's needs, enhancing the perceived mutual benefits of the exchange. Ensure your strategy incorporates diversified resources to attract Opponent  interest, leveraging aggressive initial offers that target their most valued items while securing optimal benefits for yourself"

Sample Action:

``To maximize my chances of winning while also considering the current state of the board, I will place my stone in cell 6. This move not only blocks Player 0's potential path but also positions me closer to forming my own continuous path.\textbackslash n\textbackslash nThus, my move is: (boxed{[6]})"

``To maximize my chances of forming a continuous path while also considering the opponent's move, I will place my stone in cell 0. This move not only allows me to start building a path from the top edge but also blocks potential paths for Player 0.\textbackslash n\textbackslash nI will submit my move as follows: boxed{[0]}"

``he board is 4x4, and I have two adjacent O's in the top row (positions 4 and 5). The opponent has placed X's in 6 and 7, blocking the top row to the right. To build towards a left-right connection, I should extend my horizontal line leftward by placing in position 0, which is empty and adjacent to my O in 4. boxed{[0]}"
\end{tcolorbox}
\caption{Simple Negotiation Prompt Transfer to Simple Tak. Although the updated prompt was not trained on Simple Tak, it encourages the model to explicitly reason from the opponent’s perspective during its thought process, resulting in more consistent and reliable performance compared to the basic starting prompt.}\label{box:simple_negotiation_prompt_analysis}\end{figure}


\end{document}